\documentclass[hidelinks,11pt,letterpaper]{article}
\usepackage[sectionbib,square]{natbib}
\usepackage[margin=1in]{geometry}
\usepackage{inconsolata}
\usepackage{tabularx}
\usepackage[T1]{fontenc}
\usepackage[utf8]{inputenc}
\usepackage{setspace}
\usepackage{microtype}
\usepackage{amsfonts,amsmath,amssymb,amsthm}
\usepackage{fontawesome}
\usepackage{booktabs}
\usepackage{subcaption}
\usepackage{wrapfig}
\usepackage{multicol}
\usepackage{pifont}
\usepackage[usenames,dvipsnames]{xcolor}
\usepackage[font={small,color=black,onehalfspacing}, textfont={bf}, labelfont={bf}]{caption}
\usepackage{graphicx}
\usepackage[export]{adjustbox}
\usepackage{csquotes}
\usepackage{xurl}
\usepackage{mwe}
\usepackage{enumitem}
\usepackage{titlesec}
\usepackage{palatino}
\usepackage{lipsum}
\usepackage{hyperref}       %
\usepackage{nicefrac}       %
\usepackage[english]{babel}
\usepackage{multirow}

\usepackage{bbm}

\usepackage{algorithm}
\usepackage{algpseudocode}
\usepackage{tabularx}

\usepackage[defaultlines=2,all]{nowidow} %added by Eva, to avoid orphans and widows

\usepackage{tikz}
\usetikzlibrary{shapes, positioning, shadows.blur, arrows.meta}

\titleformat{\subsubsection}[runin]% runin puts it in the same paragraph
       {\normalfont\bfseries}% formatting commands to apply to the whole heading
       {\thesubsubsection}% the label and number
       {0.5em}% space between label/number and subsection title
       {}% formatting commands applied just to subsection title
       [.]% punctuation or other commands following subsection title
\titlespacing*{\subsubsection}{\parindent}{*0}{0.25em}

\newenvironment{sciabstract}{%
\begin{quote}\small\onehalfspacing}
{\end{quote}}

\title{Learning When to Quit in Sales Conversations}
\author{%
  Emaad Manzoor$,^1$\footnote{Emaad Manzoor (emaadmanzoor@cornell.edu) is the corresponding author. He is an Assistant Professor of Marketing at Cornell University. Eva Ascarza is the Jakurski Family Associate Professor of Business Administration at Harvard Business School. Oded Netzer is the Arthur J. Samberg Professor of Business at Columbia Business School.
  We thank seminar participants at Northwestern University, Purdue University, Columbia University, the University of Texas at Austin, and the University of Southern California, in addition to participants at the New Data for Consumer Insights Conference (especially our discussant Sanjog Misra), the Customer Intelligence Lab at Harvard, and the 2025 ISMS Marketing Science conference for their feedback. We also thank an anonymous firm for providing the data, and are grateful to Javier Serra de Paz and Daniel Domínguez Mosquerra at the firm in particular. Open-source software is available at \url{stoppingagents.com}.}~~ Eva Ascarza,$^2$~ Oded Netzer,$^3$ \\
  \vspace*{\baselineskip} \\
  \small $^1$ SC Johnson College of Business, Cornell University \\
  \small $^2$ Harvard Business School, Harvard University \\
  \small $^3$ Columbia Business School, Columbia University 
}

\date{\vspace*{0.5\baselineskip}\today}

\begin{document}

\maketitle
\thispagestyle{empty}

%\onehalfspacing

\begin{center}
  % \bf\textit{Preliminary and Incomplete - Please do not distribute}

% \vspace{1cm}

\textbf{Abstract}
\end{center}
\begin{sciabstract}
  Salespeople frequently face the dynamic screening decision of whether to persist in a conversation or abandon it to pursue the next lead. Yet, little is known about how these decisions are made, whether they are efficient, or how to improve them. We study these decisions in the context of high-volume outbound sales where leads are ample, but time is scarce and failure is common. We formalize the dynamic screening decision as an optimal stopping problem and develop a generative language model-based sequential decision agent --- a \textit{stopping agent} --- that learns whether and when to quit conversations by imitating a retrospectively-inferred optimal stopping policy. Our approach handles high-dimensional textual states, scales to large language models, and works with both open-source and proprietary language models. When applied to calls from a large European telecommunications firm, our stopping agent reduces the time spent on failed calls by 54\% while preserving nearly all sales; reallocating the time saved increases expected sales by up to 37\%. Upon examining the linguistic cues that drive salespeople's quitting decisions, we find that they tend to overweight a few salient expressions of consumer disinterest and mispredict call failure risk, suggesting cognitive bounds on their ability to make real-time conversational decisions. Our findings highlight the potential of artificial intelligence algorithms to correct cognitively-bounded human decisions and improve salesforce efficiency.
\end{sciabstract}

\begin{quote}\small\textit{Keywords:} salesforce management, telemarketing, agentic AI, text analysis, conversation data\end{quote}

% {\let\thefootnote\relax\footnotetext{}}

\clearpage

\doublespacing

\pagenumbering{arabic}

\section{Introduction}

%SALES, esp. high-volume
Sales activities constitute over 5\% of the U.S.\ GDP and employ  10\% of the U.S. labor force \citep{misra2019selling}. Despite their economic significance, sales operations remain chronically time-inefficient: salespeople frequently spend valuable time on conversations that are unlikely to succeed \citep{dixon2022indecision}. This inefficiency is especially pronounced in high-volume settings like outbound call center sales\,---\,a \$97 billion global industry\, \citep{grandview2025callcenter} ---\,where the supply of leads is ample, but salespeoples' time is limited and failure is common.

%PRIOR RESEARCH and OUR FOCUS
Much of the academic and managerial attention has centered on how to motivate selling effort. In contrast, little is known about the decision of whether and when to quit a sales call that is unlikely to succeed, formally known as the dynamic qualification problem. Given the prevalence of failure in high-volume sales, even marginal improvements in qualification efficiency can yield substantial time savings.
%CONTEXT and early EVIDENCE
Indeed, in our empirical setting of an outbound sales campaign at a European telecommunications firm, calls that failed to end in a sale lasted almost 3 minutes on average. But just 1 minute into the call, a fine-tuned large language model (LLM) can predict eventual failure to sell quite accurately. Quitting these predictably-risky calls 1 minute in could have saved salespeople 55 hours without sacrificing any sales (further details in Section \ref{sec:data}).

%PREDICTION TO PRESCRIPTION
Yet, recognizing that a call is likely to fail is only part of the managerial challenge. The core decision is not just \emph{how} to predict failure, but \emph{when} to act on that prediction \citep{ascarza2021you}. To that end, we formalize dynamic qualification as an optimal stopping problem \citep{shiryaev1978optimal}, in which the algorithmic decision maker must trade off the immediate benefit of quitting against the option value of continuing to gather information at every instant. Addressing this decision problem requires more than accurate predictions: it requires a principled framework for translating evolving conversational signals into timely, high-stakes actions that maximize cumulative payoffs.

%SOLUTION
To address this decision problem, we propose a \textit{stopping agent}: a generative artificial intelligence agent that silently observes an ongoing conversation and decides whether and when to quit the conversation in real-time.\footnote{Equivalently, the stopping agent can advise the salesperson to quit, instead of deciding directly.} Although our stopping agent is a generative language model, it never interacts with the prospect. Instead, our stopping agent generates sequential quitting \textit{decisions} conditional on the observed call transcript text to maximize the expected cumulative payoff.
% , and the the salesperson interacts with the prospects and quits the call. 

By parameterizing the stopping policy with a pretrained generative LLM, our stopping agent leverages the state-of-the-art natural language understanding capabilities of such models while also overcoming the inability of dynamic programming to handle high-dimensional states (i.e., call transcripts).
However, training language models to become decision policies is challenging, and standard reinforcement learning approaches suffer from training instability, hyperparameter sensitivity, and a lack of computational scalability \citep{engstrom2020implementation,ahmadian2024back}.

We address this challenge by formulating the problem of stopping policy estimation as an \emph{imitation learning} problem. We show that optimal quitting decisions can be inferred from historical sales conversations by potentially suboptimal salespeople. We then fine-tune an LLM to generate these inferred optimal decisions given the transcript. In contrast with reinforcement learning approaches, our imitation learning approach is stable, robust to hyperparameters, scales to billion-parameter LLMs, and is compatible with proprietary LLMs only accessible via restrictive APIs.
% Thus, the objective of our LLM-based stopping agent is not to recommend the best persuasive language for the call or to continue the call beyond its observed length but rather to use the observed conversation during the call to devise an optimal stopping rule for the salesperson to terminate the call and allocate the time to more promising calls. 

%EMPIRICAL PERFORMANCE
We build a GPT-4.1 stopping agent and apply it to a dataset of 11,627 outbound sales calls collected over one month during a cross-selling campaign at our partner firm. When allowed to quit 60 or 90 seconds into a call, our stopping agent retains nearly all sales (130 out of the 132 sales observed in the held-out calls) while reducing the total call time by 36\%. Reallocating the time saved to new calls would increase expected sales by 33\%. A more `aggressive' variant of our stopping agent, which is additionally allowed to quit 30 seconds into the call, increases expected sales by 37\% by reducing the total call time by 54\%. Model training and inference costs totaled only \$150, showcasing the cost-effectiveness of our approach.

% DIAGNOSTIC ANALYSIS
We further explore systematic patterns in salespeople's quitting decisions and find converging evidence that salespeople are cognitively constrained and struggle to predict eventual call outcomes.
Specifically, we find that salespeople seem to under-react to early linguistic indicators of eventual call failure, and that their quitting decisions appear to rely on simple decision rules that overweight a few salient phrases. The phrase most predictive of salespeople quitting by far\,---\,\textit{``no me interesa''} (``I'm not interested'')\,---\,rarely appears early in the call, suggesting that salespeople waste valuable time due to waiting for this salient expression of the prospect's disinterest.

Our stopping agent, in contrast, identifies subtle linguistic indicators of a lack of conversational progress to quit earlier than salespeople. Further, and unlike salespeople, our stopping agent exhibits dynamic variation in its quitting strategy, initially focusing on whether the salesperson is speaking to the right person, then progressing to whether the prospect is interested, and finally to whether the prospect already has alternatives to the salesperson's offering.
% Further, our stopping agent exhibits dynamic variation in its quitting strategy. It initially quits based on whether the salesperson is talking to the right person, then based on the prospect's interest, The stopping agents tend to be more dynamic in its interpenetration of call progression. It quit early on (30 sec. into the call) based on linguistic indicators that signal whether it is talking to the right person. At 60 seconds, it assesses the prospect level of interest, and at 90 seconds it explore whether the prospect has reasons to adopt such as having an alternative option. 

We also examine how salespeople's quitting decisions respond to a plausible shifter of their opportunity cost of time: the proximity of the call to the end of the shift. Calls made near the end of the shift are indeed shorter on average than those made earlier, consistent with higher opportunity costs of time. Under higher time costs, a rational salesperson would selectively shorten calls that are more likely to fail. However, we find that salespeople shorten calls throughout the predicted failure risk distribution, suggesting an inability to accurately predict call failure risk.
%  Our analysis suggests that simply encouraging salespeople to shorten calls may be insufficient to improve their quitting decisions, and reinforces the need for algorithmic decision support.

%CONTRIBUTIONS
Our research makes three key contributions, spanning substantive, methodological, and managerial dimensions. \text{Substantively}, we formalize  dynamic qualification as an optimal stopping problem and propose a sequential decision-making algorithm using large language models (LLMs) to optimize quitting decisions in live sales conversations. Our approach is the first to enable proactive, data-driven termination of sales calls that balances time costs against sales potential. We show that this approach yields substantial expected sales gains and outperforms several benchmarks.

\text{Methodologically}, we develop a stopping agent that functions as a language model-based policy for optimal stopping with high-dimensional textual states. In doing so, we extend the natural language processing literature on detecting adverse conversational outcomes \citep{zhang2018conversations} to optimally stopping conversations before such outcomes occur. Our work also presents the first imitation learning solution to optimal stopping with high-dimensional states. Unlike state-of-the-art reinforcement learning approaches \citep{damera2023deep}, our approach is stable, robust to hyperparameters, and scales to billion-parameter LLMs.

\text{Managerially}, we offer a practical and cost-effective approach to improve the efficiency of high-volume outbound sales campaigns. Our stopping agent can be implemented using readily available call transcripts and deployed at a low cost (e.g., \$100-\$150 per month), even when working with proprietary language models. Moreover, our diagnostic analysis suggests that salespeople are cognitively bounded and struggle to predict call outcomes in real-time, underscoring the need for algorithmic decision support tools to alleviate their cognitive constraints.

%REST OF PAPER
The rest of our paper is organized as follows. Section~\ref{sec:related_work} reviews related literature. Section~\ref{sec:optimalstopping} presents our generative language agent for optimal stopping. Section~\ref{sec:application} applies our method to call center sales and evaluates its performance. Section~\ref{sec:diagnosing_decisions} explores drivers of salespeople's suboptimal decisions. Section~\ref{sec:closing} concludes, discusses limitations, and proposes directions for future research.

\section{Related Work}
\label{sec:related_work}

This paper studies the dynamic screening decision of whether and when a salesperson should disqualify a prospect and end the ongoing sales conversation. While qualification is recognized as a core component of the selling process \citep{misra2019selling}, dynamic qualification decisions \textit{within} sales conversations have received little attention.

% salesforce effort management
The empirical salesforce management literature is primarily focused on mechanisms to motivate selling effort along various dimensions \citep{misra2011structural,chung2014bonuses,daljord2016homogeneous,kim2019salespeople,kim2022structural,bommaraju2025multi} (we delegate to \cite{misra2019selling} for a thorough survey). Our research broadens the scope of salesforce management research by examining how effort can be optimally withheld through disqualification and reallocated to more promising prospects. Rather than designing a salesforce compensation scheme to motivate optimal disqualification, we propose an algorithmic solution to assist salespeople with optimal conversation stopping.

% salespeople decision-making
Our work contributes to the growing literature on improving salesperson decision-making using artificial intelligence algorithms. In this literature, \cite{chakraborty2025can} propose algorithms to improve salesforce recruitment, \cite{karlinsky2024automating} propose algorithms to guide pricing, \cite{hu2024zero} propose algorithms to match salespeople with the right prospects, and \cite{reeder2024improving} use LLMs to forecast sales revenue from CRM activity logs. Our approach complements this research, and goes beyond prediction to optimize sequential decisions.
More broadly, our work contributes to the literature on LLMs as collaborators \citep{arora2025ai}.

% optimal stopping, dynamic discrete choice
Methodologically, we propose an imitation learning approach to optimal stopping with high-dimensional textual or conversational states, which enables using language models as policies. 
\cite{damera2023deep} propose a policy gradients (i.e., reinforcement learning) approach to optimal stopping with high-dimensional non-textual states and recurrent neural network (RNN) policies. As we demonstrate  in Section~\ref{sec:application}, their approach underperforms in our setting.

A broader stream of research applies reinforcement learning to align LLMs with human preferences, including methods such as Proximal Policy Optimization (PPO) \citep{schulman2017proximal} and Group Relative Policy Optimization (GRPO) \citep{shao2402deepseekmath}. These approaches assume environments where the language model controls the next state via text generation, whereas the state dynamics in our setting are externally governed by the customer–salesperson interaction. Hence, methods like PPO and GRPO are not directly applicable in our settings.

Our work also relates to the economics and marketing literature on optimal stopping and dynamic discrete choice \citep{rust1987optimal}. \cite{hui2008modeling} model DVD preorders as optimal stopping decisions, \cite{yoganarasimhan2013value} models bid selection in auctions for freelance projects as optimal stopping rules, and the literature on consumer search models the search process as an optimal stopping problem \citep{zwick2003consumer,branco2016too,guo2022strategic}.
More recently, \cite{kang2025empirical} leverage the equivalence between dynamic discrete choice modeling and inverse reinforcement learning to propose a gradient-based reward function estimation approach, and \cite{barzegary2025recursive} propose reducing the state space's dimensionality via recursive partitioning.

The aforementioned literature focuses on estimating dynamic discrete choice models given observed decisions (i.e., estimating structural primitives) and simulating counterfactuals under the assumption that decision-makers behave optimally. In our work, we do not assume that salespeople behave optimally. Instead, we provide a prescriptive (algorithmic) approach to improve potentially-suboptimal stopping decisions. Indeed, we find evidence that salespeople in our setting deviate systematically from optimal stopping behavior.

% Agentic AI ReACT, SayCAN, other work in robotics and embodied LLMs.
Our work also relates to the literature on agentic artificial intelligence, wherein LLMs generate actions in addition to natural language. Methods in this literature rely both on careful prompt engineering (e.g., \citep{meta2022cicero,park2023generative,yao2023react,kim2023language,yang2024swe} and on explicitly training models to act autonomously (e.g., \citep{chen2021decision,ahn2022can,schick2023toolformer}). Our work extends this research by building an artificial intelligence agent for optimal conversation stopping and evaluating it on call center sales conversations from the field.

% \item Analyzing human decision-making errors 
Finally, our work relates to research in behavioral economics using machine learning to evaluate the quality of human decisions \citep{kleinberg2018human,mullainathan2022diagnosing,rambachan2024identifying}.
These studies test for screening errors in settings where decision-makers have a one-time choice, such as whether to detain a defendant or administer a medical test, and assess deviations from an implicit threshold rule. We contribute to this literature with an examination of salespeople's dynamic qualification decisions under time pressure. 

\clearpage

%\vspace{-20mm}
\section{Optimal Conversation Stopping with Generative Language Agents}
\label{sec:optimalstopping}

To support salespeople's dynamic qualification decisions, we develop an algorithmic agent that observes the live conversation and decides whether and when to quit. We build a \emph{language agent}: a language model that generates decisions to optimize a long-term managerial objective.
Unlike traditional conversational agents, our language agent does not interact with the prospect. Instead, it observes the conversation between the salesperson and the prospect and optimally quits (or advises quitting) to maximize expected profits. We refer to such language agents as \emph{stopping agents}.

\subsection{Problem Definition}
\label{sec:problem_definition}
Since conversations evolve sequentially, quitting decisions are inherently dynamic. Each moment spent on a call generates information about whether it is likely to end in a sale, but also incurs an opportunity cost: the time could have been allocated to a different prospect. This trade-off motivates formulating qualification as an \emph{optimal stopping problem} with textual states.
We adapt the discrete-time finite-horizon optimal stopping formulation of \cite{shiryaev1978optimal} to our setting.

Formally, at each period~$t\in\{1, \dots, T\}$, the state $s_t \in \mathcal{S}$ consists of the verbatim transcript of the conversation up to period $t$. A policy $\pi_\theta(a_t|s_t)$, parameterized by $\theta$, observes $s_t$ and selects an action $a_t \in \mathcal{A} = \{\textsf{wait}, \textsf{quit}\}$ at each time $t$. If the policy chooses \textsf{wait}, it receives a waiting reward $w_t$, and the process transitions to the next state according to the (unknown) conversational dynamics $\mathcal{P}(s_{t+1} | s_t, a_t)$.\footnote{Importantly, \textsf{wait} implies the policy \textit{doing nothing}; the conversation proceeds fully unaffected. Formally, $\mathcal{P}(s_{t+1} | s_t, a_t=\textsf{wait})=\mathcal{P}(s_{t+1} | s_t)$, and $\mathcal{P}(s_{t+1}=\textsf{end state} | s_t, a_t=\textsf{quit})=1$. The end state does not require the salesperson to quit immediately, it only implies that no further actions by the stopping policy are permitted, and is compatible with the salesperson ending the conversation naturally. This structure distinguishes optimal stopping problems from general sequential decision problems.
Our key innovation is to leverage this distinction for efficient policy estimation (Section \ref{sec:proposed_approach}).
Note that our stopping policy cannot advise continuation if the salesperson decides to quit, nor can it advise the salesperson what to say; we delegate these extensions of the action space to future work.
%However, the tradeoff of adopting the canonical optimal stopping formulation is that our stopping agent cannot enforce or advise continuation if the salesperson decides to quit. Despite this tradeoff, we demonstrate substantial efficiency gains in Section \ref{sec:application}.
} 
If the policy chooses \textsf{quit}, it receives a terminal reward $q_t$ and the process terminates.
We impose $a_{T} = \textsf{quit}$ and set $T$ to the conversation duration (i.e., a dummy terminal period), which enables denoting policies that never quit as policies that quit at $T$ without additional notation.

%\footnote{Note that the next state depends only on the conversation history and not on the previous action. This property is characteristic of optimal stopping problems but does not generally hold for broader classes of sequential decision-making tasks.}  

Let $\tau \in \{1, \dots, T\}$ denote the stopping time in a conversation induced by a policy $\pi_\theta$, defined as the earliest time at which the action is \textsf{quit}:
\begin{align*}
\tau = \min\{t \in \{1, \dots, T\} : a_t = \textsf{quit}\}.
\end{align*}
The objective is to find a policy $\pi_\theta$ that maximizes the expected cumulative reward $J(\theta)$, defined as:
\begin{align}
J(\theta)
&= 
\mathbb{E}_{\pi_\theta} \left[
\textstyle\sum_{t=1}^{\tau-1} w_t + q_\tau
\right], 
\label{eq:expected_cumulative_reward}
\end{align}
where the expectation is taken over the distribution of state-action trajectories induced by $\pi_\theta$ and the transition dynamics $\mathcal{P}(s_{t+1} | s_t, a_t)$ over all calls.

The waiting and quitting rewards $w_t$ and $q_\tau$ are exogenously specified by the firm to reflect its operational costs and demand-side conditions. For example (as we show in Section \ref{sec:application}), the firm can set $w_t$ based on the opportunity cost of time, such that the cost of waiting equals the expected revenue from spending that time on another call. Similarly, $q_\tau$ can be defined as: (a) zero if the stopping agent quits before the salesperson makes a sale, and (b) equal to the profits a successful sale generates if the stopping agent does not quit before the salesperson makes the sale.

\subsection{Algorithmic Challenges in Solving the Optimal Stopping Problem}\label{sec:challenges}

A natural starting point for solving optimal stopping problems is dynamic programming \citep{bellman1966dynamic}. In principle, the optimal value functions $V^*_t(s)= \max_\theta ~\mathbb{E}_{\pi_\theta} \left[\sum_{t=1}^{\tau-1} w_t + q_\tau \,\Big|\, s_1 = s\right]$
satisfy Bellman's optimality conditions and can be computed recursively via backward induction:
\begin{align}
V^*_t(s) = \max\left(q_t,\, w_t + \mathbb{E}[V^*_{t+1}(s_{t+1})|s_t=s, a_t=\textsf{wait}]\right), \quad V^*_T(s) = q_T.
\label{eq:bellman_equation}
\end{align}
The optimal policy at time $t$ selects \textsf{quit} if and only if $q_t > w_t + \mathbb{E}[V^*_{t+1}(s_{t+1}) \mid s_t,a_t=\textsf{wait}]$.\footnote{While one may consider a model-based policy that estimates $\mathbb{E}[V^*_{t+1}(s_{t+1}) \mid s_t,a_t=\textsf{wait}]$, this requires an accurate model of $\mathcal{P}(s_{t+1} | s_t)$, which is challenging to estimate in real-world settings. Our approach does not require $\mathcal{P}(s_{t+1} | s_t)$.} % TODO: move into new challenge

While the formulation above is conceptually straightforward, estimating the optimal stopping policy in our setting\,---\,conversational sales\,---\,presents two key challenges. These stem from the high dimensionality of textual data and the practical issues of reinforcement learning (RL) when applied to estimate large language model (LLM) policies.

\textbf{Challenge 1: Textual states and the curse of dimensionality.} 
The state at each time $t$ is a growing text sequence (i.e., the transcript of the conversation up to that point). If $\mathcal{V}$ is the vocabulary and $L$ is the maximum transcript length, the size of the state space is of the order $|\mathcal{S}| \approx $$|\mathcal{V}|^L$. This exponential growth renders dynamic programming computationally infeasible and prevents the value function from being stored or calculated in closed form. This curse of dimensionality also affects other state-enumeration approaches, such as $Q$-learning \citep{watkins1989learning}.

\textbf{Challenge 2: Practical issues with deep reinforcement learning of large language model policies.}
When enumerating a large state space is infeasible, the standard remedy is to estimate a parameterized policy $\pi_\theta(\cdot)$. We therefore parameterize our stopping policy with a pretrained generative LLM to leverage the state-of-the-art natural language understanding abilities of such models \citep{kaplan2020scaling}. Estimating policies parameterized by LLMs is, in principle, a natural fit for deep reinforcement learning methods, which have recently been used in several marketing applications (\citep[e.g.,][]{liu2023dynamic, ko2024target, ma2025dynamic}).

In practice, however, coupling deep reinforcement learning with LLMs introduces well-known difficulties\footnote{Recent methods to align LLMs with human preferences \citep{schulman2017proximal,shao2402deepseekmath,lambert2024t} assume that the next state is formed by appending the generated token to the previous state. In our setting, this assumption does not hold: the next state is determined by the conversation between the customer and the salesperson.} \citep{henderson2018deep, engstrom2020implementation,ahmadian2024back}, notably: (1) unstable optimization performance that is highly sensitive to hyperparameters and implementation details; and (2) substantially higher computational costs than supervised learning.
These practical difficulties have renewed interest in supervised learning-based approaches to policy estimation as simpler, robust, and computationally scalable alternatives \citep{foster2024is}.
% In section (\ref{sec:baselines}) we empirically compare our approach with such reinforcement learning methods.

\subsection{Proposed Method: Imitation Learning to Quit}
\label{sec:proposed_approach}

Given the aforementioned challenges, we propose a solution based on \emph{imitation learning} \citep{pomerleau1988alvinn}. Imitation learning is a form of supervised learning, and has been successfully used to train decision-making policies in applications ranging from autonomous helicopters \citep{abbeel2004apprenticeship} to self-driving cars \citep{bansal2018chauffeurnet}. In essence, imitation learning trains a policy to mimic the actions of an expert policy (such as a human driver) by learning from a dataset of optimal state-action trajectories generated by the expert.

Formally, given a dataset of state-action trajectories $\mathcal{D} = \{(s_1^i, a_1^i), \dots, (s_t^i, a_t^i)\}_{i=1}^N$ consisting of state-action pairs from an ``expert'' policy that maximizes the expected cumulative reward in Equation~\eqref{eq:expected_cumulative_reward}, imitation learning seeks a policy $\pi_{\hat{\theta}}(a|s)$ that minimizes the expected action mismatch:
\[
\hat{\theta} = \arg \min_\theta ~\mathbb{E}\!\left[\mathbbm{1}\{\pi_{\theta}(\cdot|s_t^i)\neq a_t^i\}\right],
\]
that is, imitation learning finds a policy that replicates the behavior of this ``expert'' policy and, in doing so, maximizes the expected cumulative reward.
%\footnote{\cite{kumar2022should} prove that, given optimal state-action trajectories, imitation learning and reinforcement learning have identical worst case error bounds. \cite{foster2024is} derive additional desirable properties of imitation learning.}

A key requirement of imitation learning is a dataset of state-action trajectories $\mathcal{D}$ from a policy that maximizes the expected cumulative reward in Equation \eqref{eq:expected_cumulative_reward},
% That is, a dataset of a large set of optimal decisions that can serve as training for the imitation learning estimation. 
which might appear prohibitive. In settings such as ours, we cannot assume that salespeople behave optimally (in fact, we show evidence of their suboptimality in Section \ref{sec:diagnosing_decisions}). Our key insight is that \textit{despite} their suboptimality, for the purpose of optimizing the objective in Equation \eqref{eq:expected_cumulative_reward}, $\mathcal{D}$ can be constructed using historical conversations. Specifically, we infer the cumulative reward-maximizing action $a_t$ for each transcript prefix $s_t$ in a historical conversation by calculating the quitting and waiting rewards given the actual conversation outcome. We then train an LLM to imitate (i.e., by generating) the cumulative reward-maximizing actions given the state. Critically, this reduces the original objective in Equation~\eqref{eq:expected_cumulative_reward} to the scalable and cost-effective task of fine-tuning an LLM.

Our approach bypasses both the intractability of dynamic programming over high-dimensional textual state spaces (Challenge 1) and the instability and hyperparameter sensitivity of reinforcement learning with LLMs (Challenge 2). Moreover, by framing policy learning as language model fine-tuning, our method inherits the scalability and tooling of modern language model fine-tuning pipelines. Further, our approach is readily compatible with proprietary language models that disallow custom loss functions, and is extensible for use with multi-modal language models. 

\begin{figure*}[t]
  \centering
  % \vspace{-5mm}
  \begin{tikzpicture}
    \centering
    % Define a prompt‐box style with a soft blurred shadow
    \tikzset{
      promptbox/.style={
        draw=black,                    
        fill=yellow!15,                  
        % rounded corners=3pt, 
        shadow blur radius=2pt,
        shadow blur steps=5,          
        inner sep=8pt,                
        text width=5cm,                
        align=left,      
        shadow xshift=2pt,             
        shadow yshift=-2pt             
      },
      responsebox/.style={
        draw=black,
        fill=blue!25,
        % rounded corners=3pt,
        inner sep=6pt,
        text width=1cm,
        align=center,
        shadow blur steps=5,
        shadow blur radius=2pt,
        shadow xshift=2pt,
        shadow yshift=-2pt
      },
      callbar/.style={
        draw=black,
        % rounded corners=2pt,
        minimum height=5mm,
        inner sep=0pt
      },
      oainode/.style={
        % draw=black,
        % fill=white,
        % rounded corners=4pt,
        % inner sep=8pt,
        minimum width=3cm,
        minimum height=3cm,
        align=center,
        % drop shadow
      }
    }

    %
    % 1) Dataset of calls on the left
    %
    % Label for dataset
    \node[font=\bfseries] (dslabel) at (0,2.75) {\footnotesize Conversation Dataset};

    \node[callbar, fill=green!10] (call1) at (0,2) [minimum width=2cm] {\footnotesize \color{gray} \textit{perfecto \dots}};
    \node[callbar, fill=pink!25]   (call2) at (0,1.4) [minimum width=3.5cm] {\footnotesize \color{gray} \textit{que no se si ya has \dots}};
    \node[callbar, fill=green!10] (call3) at (0,0.8) [minimum width=4cm] {\footnotesize \color{gray} \textit{hola te llamamos de tu \dots}};
    \node[callbar, fill=pink!25]   (call4) at (0,0.2) [minimum width=4cm] {\footnotesize \color{gray} \textit{en agradecimiento  \dots}};
    \node[callbar, fill=green!10] (call5) at (0,-0.4) [minimum width=2cm] {\footnotesize \color{gray} \textit{encantada \dots}};
    \node[callbar, fill=pink!25]   (call4) at (0,-1) [minimum width=4cm] {\footnotesize \color{gray} \textit{no tengo mas \dots}};
    \node[callbar, fill=pink!25]   (call4) at (0,-1.6) [minimum width=1.5cm] {\footnotesize \color{gray} \textit{que \dots}};
    \node[callbar, fill=pink!25]   (call4) at (0,-2.2) [minimum width=3cm] {\footnotesize \color{gray} \textit{comentarte \dots}};
    \node[callbar, fill=pink!25]   (call4) at (0,-2.8) [minimum width=2cm] {\footnotesize \color{gray} \textit{electrica \dots}};
    \node[below=1mm of call4.south] {\color{gray}\footnotesize Red$\equiv$Failed};
    \node[below=4mm of call4.south] {\color{gray}\footnotesize Green$\equiv$Succeeded};
  
    % Draw three overlapping prompt boxes
    \node[promptbox] (card1) at (6,0) {
      \footnotesize
      Estos son los primeros \textit{t} segundos de la conversación entre el agente de ventas y el consumidor:\\[1.2em]

      \textit{Orador 1: perfecto perfecto}\\
      \textit{Orador 2: pues eso ahorros \dots}\\
      \textit{\dots}\\[1.2em]

      ¿Deberíamos abandonar esta conversación ahora o esperar (responder abandonar o esperar)? 
    };
    % Label above the stack
    \node[above=0mm of card1.north] (promptlabel) {\bfseries\centering\footnotesize States $s_t$ Wrapped in Prompts};
  
    \draw[-{Stealth[length=3mm]}, thin]
    (dslabel.east) .. controls +(right:10mm) and +(-1,1) .. (promptlabel.west);

    \node[promptbox, xshift=6pt, yshift=-6pt] (card2) at (card1) {
      \footnotesize
      Estos son los primeros t segundos de la conversación entre el agente de ventas y el consumidor:\\[1.2em]

      \textit{Orador 1: perfecto perfecto}\\
      \textit{Orador 2: pues eso ahorros \dots}\\
      \textit{\dots}\\[1.2em]

      ¿Deberíamos abandonar esta conversación ahora o esperar (responder abandonar o esperar)? 
    };
    \node[promptbox, xshift=12pt, yshift=-12pt] (card3) at (card1) {
      \footnotesize
      Estos son los primeros t segundos de la conversación entre el agente de ventas y el consumidor:\\[1.2em]

      \textit{Orador 1: perfecto perfecto}\\
      \textit{Orador 2: pues eso ahorros \dots}\\
      \textit{\dots}\\[1.2em]

      ¿Deberíamos abandonar esta conversación ahora o esperar (responder abandonar o esperar)? 
    };

    \coordinate (anchor) at ([xshift=59pt]card1.south);
    % Draw three overlapping response boxes below
    \node[responsebox, below=6mm of anchor] (res1) {
      \footnotesize
      esperar
    };
    \node[responsebox, xshift=6pt, yshift=-6pt] (res2) at (res1) {
      \footnotesize
      esperar
    };
    \node[responsebox, xshift=12pt, yshift=-12pt] (res3) at (res1) {
      \footnotesize
      esperar
    };
    \node[left=5mm of res1.west] (resplabel) {\footnotesize \textbf{Expected Generation}};
    \node[below=-2mm of resplabel] {\footnotesize $\equiv$ Expert's Action $a_t$};

    % 4) OpenAI node on the far right
    \node[oainode] (openai) at (12.5,1.25) {
      \footnotesize \bfseries Generative Large\\[-1mm]
      \footnotesize \bfseries Language Model $\pi_\theta(a|s)$\\[2mm]
      \includegraphics[width=2.5cm]{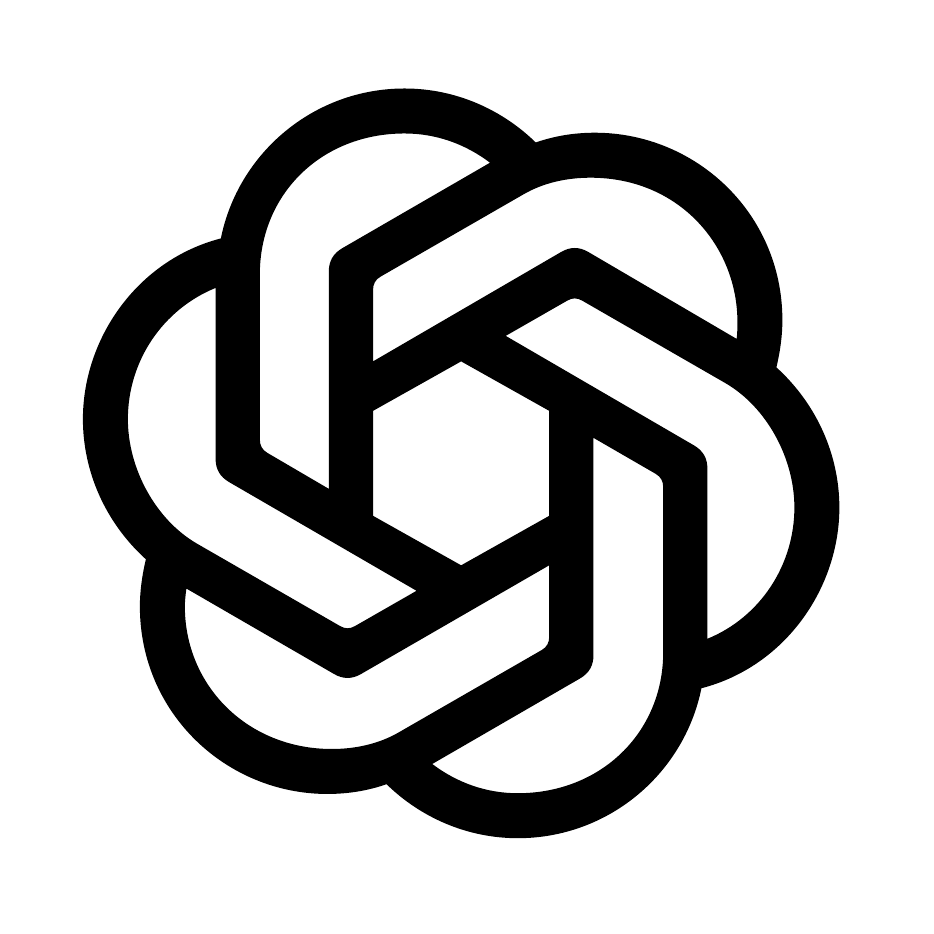}
    };
     % … (after your OpenAI node)
    \node[below=8mm of openai.south,  inner sep=12pt,align=center] (loss) {
      % \footnotesize
      \(-\log \pi_\theta\bigl(a_t | s_t\bigr)\)
      \color{gray} (log loss)
    };

    % 5) Fine-tune arrows from prompts & responses into OpenAI node
    \draw[-{Stealth[length=3mm]}, thin]
    (promptlabel.east) .. controls +(right:10mm) and +(-1,1) .. ([yshift=1.5cm]openai.west);

    \draw[-{Stealth[length=3mm]}, thin]
  (res3.east) -| (loss.south);

  \draw[-{Stealth[length=3mm]}, thin] 
  (openai.south) -- (loss.north);
  \end{tikzpicture}
  \caption{We transform conversations $\mathcal{C}$ into ``expert'' demonstrations $\mathcal{D}$ (Section~\ref{sec:expert_demonstrations}), and train an imitation learning policy by fine-tuning an LLM $\pi_\theta$ to generate the ``expert's'' action $a_t$ for each state $s_t$ wrapped in a prompt (Section~\ref{sec:finetuning}). We threshold $\pi_\theta(a|s)$ to obtain deterministic actions (Section~\ref{sec:threshold_tuning}).}
  \label{fig:proposed_approach}
  % \vspace{-1mm}
\end{figure*}
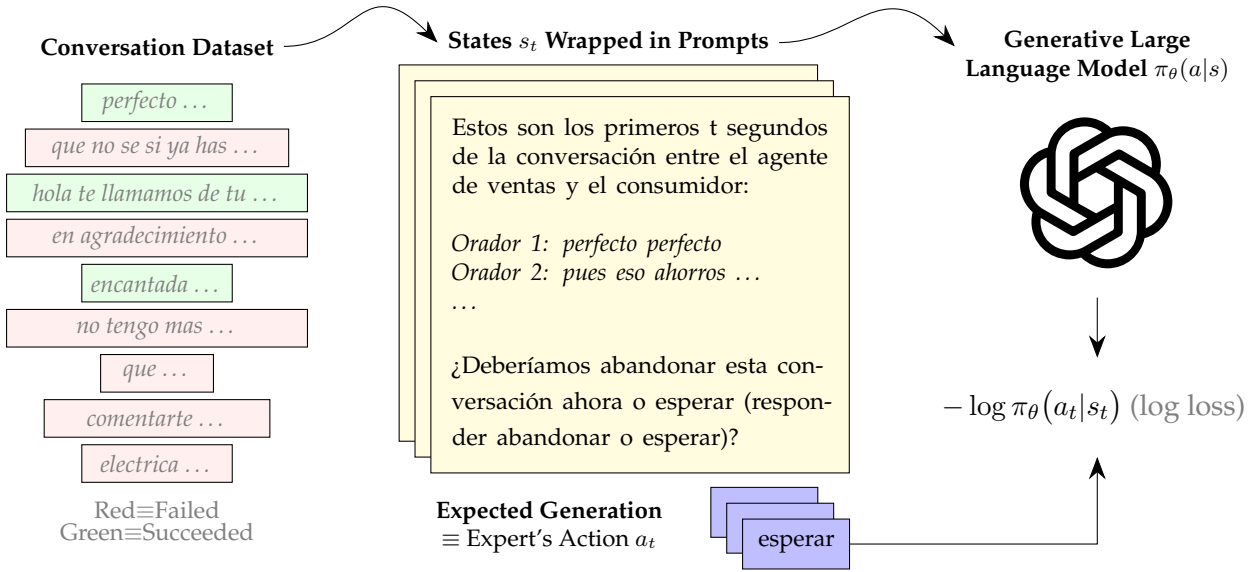

Figure~\ref{fig:proposed_approach} summarizes our approach. We now describe each of these steps in detail.

\subsubsection{Inferring Optimal State-Action Trajectories from Historical Conversations}
\label{sec:expert_demonstrations}

To construct the dataset $\mathcal{D}$ of optimal state-action trajectories using historical conversations, we calculate, for each historical conversation, the quitting time $\tau$ that would have maximized the expected cumulative reward given $w_t$ and $q_\tau$.
% For $w_t$ we use the opportunity cost of using the call time to make another call with and average expected likelihood of success, and for $q_t$ we use 1 of the sales ends in a sales the quitting decision is at $t=T$ and 0 otherwise. Given this cost structure and the observed calls, we optimize the quitting decision based on Equation (Equation). 
This induces a state-action trajectory comprised of transcripts  and the corresponding optimal actions with respect to Equation \eqref{eq:expected_cumulative_reward}, thereby representing an expert policy without needing access to optimal quitting decisions from salespeople.

The reason why this works lies in the structure of the optimal stopping problem defined in Section \ref{sec:problem_definition}. For optimal stopping problems, the next state $s_{t+1}$ of a historical conversation given the current state $s_t$ and an action $a_t$ is known: it is either the terminal state if $a_t=\textsf{quit}$, or the conversation transcript until $t+1$ if $a_t=\textsf{wait}$ (since \textsf{wait}ing implies doing nothing). This structure distinguishes optimal stopping problems from general sequential decision-making problems, and our key innovation is to leverage this structure for imitation learning.
% MOVED THIS TO EARLIER
%\footnote{We estimate stopping policies that optimize Equation \eqref{eq:expected_cumulative_reward}, but not policies that advise what to say or whether to continue the call when the salesperson decides to quit. We delegate these extensions of the action space to future work.
%However, the tradeoff of adopting the canonical optimal stopping formulation is that our stopping agent cannot enforce or advise continuation if the salesperson decides to quit. Despite this tradeoff, we demonstrate substantial efficiency gains in Section \ref{sec:application}.}
% Note that this optimal stopping problem indeed provides optimal quitting given actual call transcript but not a suggestion to continue the call beyond the observed call length or modify the call conversation, which is beyond the scope of this work.

Formally, let $\mathcal{C} = \{(s_1^j, s_2^j, \dots, s_T^j, y^j)\}_{j=1}^M$ denote a dataset of $M$ conversations, where $s_t^j$ is the transcript of conversation $j$ up to time $t$, and $y^j \in \{0, 1\}$ indicates whether the conversation ended in a sale. For each conversation $j$ and each candidate quitting time $\tau \in \{1, \dots, T\}$, we compute the cumulative reward that would have been obtained by quitting at $t=\tau$ as,
\begin{align}
R_j(\tau) = \sum_{t=1}^{\tau-1} w_t + q_\tau.
\label{eq:reward_tau}
\end{align}
Let $\tau^*_j = \arg \max_\tau R_j(\tau)$ be the quitting time that maximizes cumulative reward. The corresponding state-action trajectory consists of the transcript prefixes and optimal actions up to time $\tau^*_j$, i.e., $\{(s_t^j, a^j_t)\}_{t=1}^{\tau^*_j}$, where
$a_t^j = \textsf{wait}$ for $t < \tau^{*j}$ and $a_{\tau^{*j}}^j = \textsf{quit}$.

We define $\mathcal{D}$ as the union of these optimal state-action trajectories across all conversations: 
\begin{align*}
\mathcal{D} = \bigcup_{j=1}^M \{(s_t^j, a^j_t)\}_{t=1}^{\tau^*_j}.
\end{align*}

\paragraph{Data augmentation to enable recovering from suboptimal states.} The aforementioned procedure does not include states reached via suboptimal actions in $\mathcal{D}$.  For example, if the optimal stopping time is $\tau^*=2$ for conversation $j$, $\mathcal{D}$ will include $(s^j_1, \textsf{wait})$ and $(s^j_2, \textsf{quit})$ but will omit $(s^j_t, a^j_t)$ for all $t > \tau^*$. This exclusion poses a risk: if the policy ever reaches a suboptimal state at inference-time (i.e., by \textsf{wait}ing when the optimal action was to \textsf{quit}), it may not know how to act in this state.

We address this risk using the data augmentation strategy of \citet{pomerleau1988alvinn} in the context of autonomous vehicle control (i.e., to teach a self-driving vehicle what to do after it drives off the road). Specifically, we augment $\mathcal{D}$ to teach the agent to recover from suboptimal states by adding the suboptimal-state-optimal-action pair $(s_t^j, \textsf{quit})$ for each $t = \tau^*+1, \dots, T$ and for each conversation $j$ having optimal stopping time $\tau^*$. That is, for each conversation with an optimal stopping time of $\tau^*<T$, we include in $\mathcal{D}$ the optimal decisions for time periods after $\tau^*$.

\subsubsection{Expert Mimicry as Conditional Language Generation}
\label{sec:finetuning}
The dataset $\mathcal{D}$ comprises textual states and their corresponding optimal actions (\textsf{quit} or \textsf{wait}). Hence, we train an LLM to mimic the optimal actions by fine-tuning it to \textit{generate} the \textsf{quit} and \textsf{wait} text (i.e., tokens) conditional on the conversation transcript text up to that point.
Specifically, we estimate an LLM policy $\pi_{\theta}$ by minimizing the empirical log-loss of its state-conditioned token generation over $\mathcal{D}$:
\begin{align}
  \hat{\theta} = \arg\min_\theta~\mathbb{E}_{(s_t, a_t) \sim \mathcal{D}}\left[-\log \pi_\theta(a_t | s_t)\right].
  \label{eq:log_loss}
  \end{align}
Thus, we transform the problem of maximizing the expected cumulative reward to that of fine-tuning an LLM to generate the optimal action $a_t$ given the state $s_t$. $s_t$ may be wrapped in a prompt (our specific prompt is in Figure \ref{fig:proposed_approach}), which can include any desired context (e.g., metadata, images).
  
% \paragraph{From Probabilities to Deterministic Actions: Backward Induction Threshold Tuning}\label{sec:threshold_tuning}

\subsubsection{From Probabilities to Deterministic Actions: Backward Induction Threshold Tuning}\label{sec:threshold_tuning}
Training an LLM policy by minimizing Equation \eqref{eq:log_loss} yields a stochastic policy $\pi_{\hat{\theta}}(a|s)$ that outputs a probability distribution over actions, i.e., $\pi_{\hat{\theta}}(a \mid s) \in [0,1]$. To implement this policy in practice, we must convert its probabilistic outputs into deterministic decisions. We do so by introducing thresholds $\lambda_t \in [0,1]$ for each decision point $t = 1, \dots, T$, such that the policy selects \textsf{quit} if $\pi_{\hat{\theta}}(\textsf{quit} \mid s_t) \geq \lambda_t$ and \textsf{wait} otherwise. 

While a grid search over $\lambda_1, \dots, \lambda_T$ is a straightforward tuning approach, it scales exponentially as $O(B^T)$, where $B$ is the number of grid points. To overcome this, we propose a backward induction-style threshold tuning procedure that scales linearly in $T$. The idea is to start at $T$, where quitting is mandatory, and inductively move backwards. At each time step $t$, the threshold $\lambda_t$ is set to maximize the expected reward, accounting for both the immediate reward from quitting and the future reward from waiting at $t$ and quitting later.
Algorithm~\ref{alg:threshold-selection} summarizes the procedure.\\[-1.5em]
\begin{algorithm}[h!]
  \caption{Backward Induction Threshold Tuning}
  \label{alg:threshold-selection}
  \begin{algorithmic}[1]
  \Require Validation dataset $\mathcal{D} = \{(s^i_1, \dots, s^i_T, y^i)\}_{i=1}^N$, fine-tuned stochastic policy $\pi_{\hat{\theta}}$, horizon $T$
  % \Ensure Tuned thresholds $\{\lambda_t\}_{t=1}^T$
  \State $\lambda_T \gets 0$ \Comment{Always \textsf{quit} at $T$}
  \ForAll{$(s^i_1, \dots, s^i_T, y^i) \in \mathcal{D}$}
      \State $R_T^i \gets q_T$
  \EndFor
  \For{$t = T-1, T-2, \dots, 1$}
      \ForAll{$(s_1^i, \dots, s_T^i, y^i) \in \mathcal{D}$}
          \State $R_t^i \gets q_t$ \Comment{Reward for quitting at $t$}
          \State $\tau \gets \min\{u > t : \pi_{\hat{\theta}}(\textsf{quit} \mid s_u^i) \geq \lambda_u\}$ \Comment{Next quitting time if agent waits at $t$}
          \State $R_\tau^i \gets$ reward of quitting at $\tau$ \Comment{Known by induction for $\tau \in \{t+1, \dots, T\}$}
      \EndFor
      \State $\displaystyle
      \lambda_t \gets \arg\max_{\lambda \in [0,1]} \frac{1}{N} \sum_{i=1}^N
      \left(
      \mathbb{I}[\pi_{\hat{\theta}}(\textsf{quit} \mid s_t^i) \geq \lambda] R_t^i +
      \mathbb{I}[\pi_{\hat{\theta}}(\textsf{quit} \mid s_t^i) < \lambda] R_\tau^i
      \right)$
  \EndFor
  \State \Return $\{\lambda_t\}_{t=1}^T$
  \end{algorithmic}
\end{algorithm}

\vspace{-2em}
\subsection{Putting It All Together}
We introduce stopping agents --- generative language models that make real-time disqualification decisions in conversational sales settings. These agents solve optimal stopping problems over textual state spaces by mimicking an inferred expert policy. Our proposed approach avoids the practical limitations of reinforcement learning by relying instead on imitation learning, enabling stable and cost-effective training through standard language model fine-tuning.

Stopping agents are silent companions to salespeople: they observe the evolving transcript of a sales conversation and decide (or advise the salesperson) whether and when to terminate the call. They are practical to implement and compatible with proprietary language model APIs. They can be easily initialized with visual-, audio-, and multi-modal language models. To facilitate adoption and further research, we provide an open-source framework to build, train, and evaluate stopping agents at {\small\url{stoppingagents.com}}.\footnote{Note that while stopping agents are trained offline, they are deployed in live conversations and operate in real-time.}
Having introduced the design and training of our stopping agent, we now apply it to real-world sales data to evaluate its performance.

% ~\\[-6em]
\section{Empirical Application}
\label{sec:application}

We study an outbound sales operation at a large telecommunications firm in Europe. The firm conducts cross-selling campaigns targeting its existing mobile subscribers. All sales calls were initiated through live phone calls by commissioned salespeople.\footnote{Their compensation consists of a base pay plus a bonus across a two-shift workday with a fixed number of hours. A large number of leads are assigned to each salesperson at random (i.e., salespeople do not choose their prospect pool).} The calls follow a standardized script designed to introduce and promote complementary products and services, though salespeople may deviate from the script when needed. The dataset used in our analysis comprises first-contact calls in which mobile subscribers are offered the opportunity to switch their electricity provider.

\subsection{Data}
\label{sec:data}

The dataset covers a one-month period and includes 11,627 outbound sales calls placed by 79 different salespeople. For each call, we observe the complete anonymized conversation transcript, automatically transcribed and timestamped at the utterance-level using a speech-to-text system based on the Whisper model \citep{radford2023robust}. In addition, we observe the salesperson's identifier, the call outcome (i.e., whether the call resulted in a sale based on the consumer confirming the energy contract, hereafter call \textit{success}), and metadata such as the call start and end times.

\begin{table}[h!]
  \centering
  \small
  \caption{Descriptive statistics.}
  \label{tab:data_summary}
  \begin{tabular*}{\linewidth}{@{\extracolsep{\fill}} lrrrr}
      \toprule
       & Min & Max & Mean & Standard Dev. \\
      \midrule
      Call Success (Binary) & 0 & 1 & 0.055 & 0.228 \\
      Call Duration (s) & 60 & 3453 & 195 & 214 \\
      \qquad Failed Call Duration (s) & 60 & 3453 & 169 & 163 \\
      \qquad Successful Call Duration (s) & 79 & 3094 & 630 & 417 \\
      Salesperson Success Rate & 0 & 16.9\% & 6.6\% & 4.4\% \\
      Salesperson Call Volume & 1 & 830 & 147 & 186 \\
      \bottomrule
  \end{tabular*}
\end{table}
% success rate over days, through the day

Table~\ref{tab:data_summary} summarizes key descriptive statistics.
Only 5.5\% of calls succeeded. Failed calls accounted for 82\% (517 hours) of the total time spent on calls, with an average duration of 169 seconds. Successful calls were fewer in number but longer on average, reflecting the additional time required to transact and close the sale. The average salesperson made 147 calls of which only 6.6\% succeeded. Overall, these patterns suggest that failed calls are significantly time-consuming. Our objective is to reduce the time spent on failed calls without compromising the number of sales.

For our subsequent analysis, we randomly partition our dataset into a training set of 5,690 calls ($\sim$50\%), a validation set of 3,499 calls ($\sim$30\%), and a test (i.e., held-out) set of 2,438 calls ($\sim$20\%), stratified to ensure that the proportion of successful calls ($\sim$5.5\%) is consistent across all splits.

\textbf{Motivating evidence.} To motivate our work, we build a predictor of call failure risk using the early call transcript. This predictor does not make decisions; it only estimates failure risk at a specified time instant, and does not know when to act on these predictions (in contrast with our stopping agent). We use this predictor to study two questions: (1) do sales call transcripts contain early linguistic indicators of eventual call failure?, and (2) how do salespeople react to them? 

\begin{figure*}[h!]
  \vspace{0.5em}
  \centering
  \caption{Empirically assessing (a) how well the predicted failure risk correlates with the actual failure rate, and (b) whether and how salespeople react to early indicators of eventual call failure.}
  \begin{subfigure}[t]{0.42\textwidth}
    \centering
    % \vspace{-53mm}
    \includegraphics[width=\linewidth]{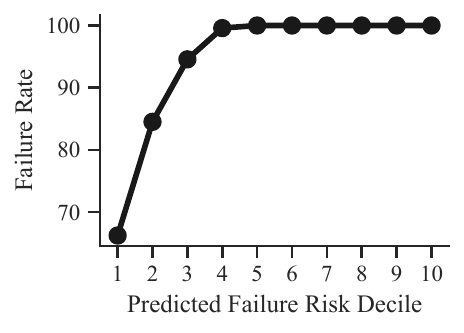}
    \caption{}
    \label{fig:risk_vs_outcome}
  \end{subfigure}
  \hspace{1em}
  \begin{subfigure}[t]{0.42\textwidth}
    \centering
    \includegraphics[width=\linewidth]{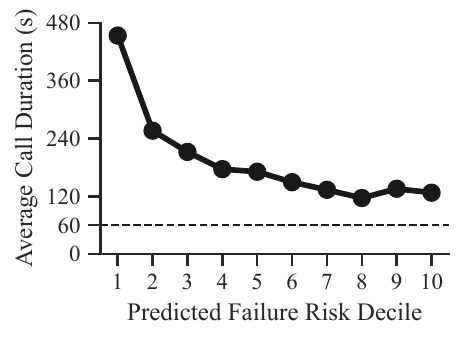}
    % \vspace{5.2mm}
    \caption{}
    \label{fig:risk_vs_duration}
  \end{subfigure}
  % \caption*{\footnotesize \textit{Note:} Failure risk of calls in the test set predicted by a GPT-4.1 classifier fine-tuned on calls in the training set.}
  \label{fig:information}
  \vspace{-1em}
\end{figure*}

Our predictor is the GPT-4.1 \citep{openai2025gpt41} large language model fine-tuned on the training set calls to predict eventual failure to sell given the first 60 seconds of the call transcript text.\footnote{Specifically, we perform supervised fine-tuning on prompt-expected response pairs using the OpenAI API. As our prompt, we use the Spanish translation of ``Here is the transcript of the first 60 seconds of a sales call between a consumer and a salesperson. \texttt{<transcript>} Will this call eventually end in a sale? (respond with `yes' or `no'):''. Our expected response is `yes' or `no' based on whether the call actually ended in a sale. We train for 3 epochs and use the validation set for early stopping. We use the probability of generating the `no' token as the predicted failure risk.\label{fn:train}} We apply this predictor to the test set calls, sort them into deciles of predicted failure risk estimated at $t=60$, and plot the actual failure rate in each predicted failure risk decile in Figure \ref{fig:risk_vs_outcome}.

Figure \ref{fig:risk_vs_outcome} shows that the failure risk predicted by GPT-4.1 given the first 60 seconds of the call transcript is quite correlated with the actual failure rate. In fact, the predictor achieves a held-out AUC (which measures predictive performance, where 100\% is best) of 94\%. Figure \ref{fig:risk_vs_outcome} thus supports the existence of early indicators of eventual call failure in the call transcript text. In the absence of such indicators, we would expect a lower out-of-sample predictive performance.

Figure \ref{fig:risk_vs_duration} further shows that salespeople spend less time on average on predictably-risky calls. The downward-sloping curve indicates that salespeople are responsive to early indicators of eventual call failure.
However, they seem to \textit{under}-react. On calls in the 6 highest predicted failure risk deciles, salespeople spent a total of 55 hours and an average of 2.3 minutes per call, though (as shown in Figure \ref{fig:risk_vs_outcome}) \textit{all} of these calls ultimately failed to end in a sale. These predictably-risky calls could have been abandoned just 60 seconds in with \emph{no loss} in sales.

The presence of early indicators of eventual call failure, in addition to salespeople's under-reaction to them, motivates our proposed decision support tool to stop sales calls early.

\subsection{Training our Stopping Agent}
\label{sec:stopping_agent_configuration}

We now describe how to train and evaluate our stopping agent using the dataset in Section \ref{sec:data}.

\paragraph{Reward configuration.}
\label{sec:reward_specification}

To configure the reward structure of our stopping agent, we specify the waiting and quitting rewards $w_t$ and $q_t$ in Equation~\eqref{eq:expected_cumulative_reward}.
We define $w_t = -c$, where $c>0$ is the opportunity cost per unit time. We set $c$ to reflect the expected value of reallocating the time spent towards initiating new calls and generating new sales, under the assumption that calls are drawn from the same distribution\footnote{This is a reasonable assumption in high-volume sales settings where the prospect supply is ample. In other settings, the reward structure can accommodate factors such as lower labor costs, improved salesforce morale, or (as noted by our partner firm) improved customer satisfaction from salespeople spending less time on calls with low-potential prospects.\label{fn:prospectsupply}} (i.e., with the same success rate and average duration):
\begin{align*}
  \qquad c = \frac{1}{\textrm{Average call duration}} \times \textrm{Success rate} \times \textrm{Time until next period}.
\end{align*}
We define the quitting reward as $q_t = b\mathbb{I}[y=1 \wedge t=T]$, where $y \in \{0,1\}$ indicates whether the call succeeded, and $b>0$ denotes the benefit of a sale. We set $b=1$ so each sale yields a unit reward. Critically, this reward is only realized if the agent chooses not to quit at any point during the call.

\paragraph{Training and evaluating the policy.} Following Section \ref{sec:proposed_approach}, we train our stopping agent by fine-tuning GPT-4.1\footnote{We use the \texttt{gpt-4.1-2025-04-14} checkpoint. However, our stopping agent is agnostic to the specific model. In Section \ref{sec:os_comparison}, we demonstrate good performance even with small open-source language models. Our training procedure is identical to that in footnote \ref{fn:train}, but uses the prompt and expected response as specified in Figure \ref{fig:proposed_approach}. We found that performance is insensitive to the exact prompt, since fine-tuning tends to eliminate its effect as a prior.} on an imitation dataset $\mathcal{D}$ inferred from calls in the training set. We tune the action thresholds on the validation set, and report all performance metrics on the held-out test set. 

\subsection{Evaluating our Stopping Agent's Performance}\label{sec:time_saved}
We start by evaluating a simple stopping agent with $T=2$ decision opportunities (ignoring the dummy terminal time period) at $t=60$ and $t=90$ seconds. We will later increase $T$ to examine the value of more frequent decision-making.
% Action prediction AUCs

\begin{figure*}[h!]
  \vspace{0.75em}
  \centering
  \caption{Total time spent (left) and total number of sales made (right) by each salesperson, both with ($y$-axis) and without ($x$-axis) our stopping agent with $T=2$ decision opportunities, at $t=60$ and at $t=90$.}
  \begin{subfigure}[b]{0.457\textwidth}
    \centering
    \includegraphics[width=\linewidth]{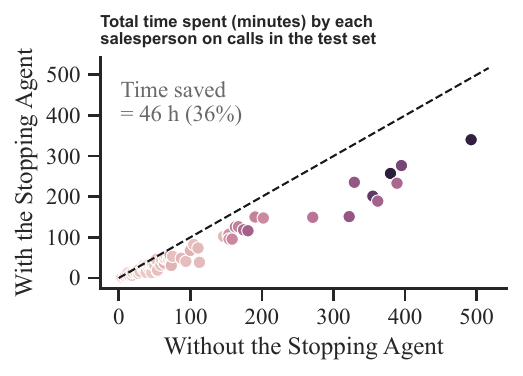}
  \end{subfigure}
  \hfill
  \begin{subfigure}[b]{0.533\textwidth}
    \centering
    \includegraphics[width=\linewidth]{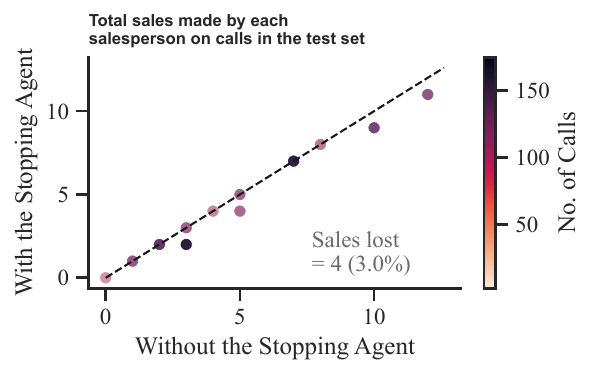}
  \end{subfigure}
  \caption*{\footnotesize \textit{Note:}
  Each point corresponds to a salesperson. The $x=y$ diagonal corresponds to no time saved and no sales lost by using the stopping agent. Stopping agent outcomes are obtained by applying it to each salesperson's test-set calls.}
  \label{fig:benchmark_time_and_sales}
  \vspace{-1em}
\end{figure*}

We begin by comparing the \emph{total time spent} and \emph{total number of sales} by each salesperson with and without our stopping agent, as shown in Figure~\ref{fig:benchmark_time_and_sales}. Each point  represents a salesperson, with color intensity indicating their call volume in the test set. Points on the $x=y$ diagonal correspond to no change in time or sales from using the stopping agent. For instance, in the left panel, points below the diagonal are salespeople who would have saved time using our stopping agent.
% Similarly, in the right panel, points below the diagonal indicate that the salesperson would have made fewer sales had they used our stopping agent.

Salespeople spent a total of 128 hours on calls in the test set. Our stopping agent reduces call time by 1.37\% to 71.6\% per salesperson, totaling 46 hours across all salespeople (a 36\% reduction in total call time). This reduction comes at the cost of 4 sales, or 3\% of the 132 sales in the test set.

To quantify the net gains from quitting earlier than salespeople, we construct a unified metric that combines time savings and observed sales. Specifically, we measure the \emph{expected} number of sales assuming that the time saved by quitting early is reallocated to initiating new calls:
\begin{align}
  \text{Expected sales} = \text{Actual sales} + \frac{\text{Time saved}}{\text{Average call duration}} \times \text{Success rate},
  \label{eq:expectedsalesgain}
\end{align}
As in footnote \ref{fn:prospectsupply}, this calculation assumes an ample supply of prospects drawn from the same distribution as the original calls (i.e., with the same average call duration and success rate).
%  In practice, this assumption is reasonable when the pool of eligible leads exceeds the number of leads actually contacted, as is the case in our empirical setting.

\begin{figure}[h!]
  \vspace{0.5em}
  \centering
  \caption{Expected number of sales by our stopping agent with $T=2$ decision opportunities at $t\in\{60,90\}$.}
  \label{fig:expected_sales_T2}
  \includegraphics[width=0.65\linewidth]{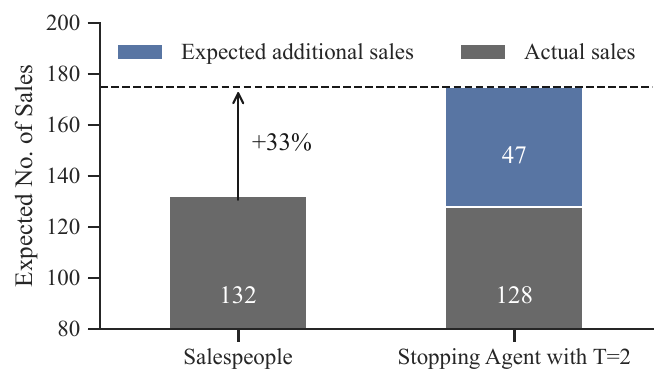}
  \caption*{\footnotesize \textit{Note:} The expected additional sales is the number of sales expected by reallocating the time saved (46 hours or 36\% saved) to making new calls drawn from the same distribution (i.e., with the same average duration and success rate).}
  \vspace{-2.5em}
\end{figure}

We visualize the actual and expected number of sales on the test set calls for salespeople and for our stopping agent in Figure~\ref{fig:expected_sales_T2}. With only $T=2$ decision opportunities at $t=60$ and $t=90$ seconds, our stopping agent delivers a 33\% increase in expected sales (from 132 to 175 sales). Of these, 128 are from the original calls and 47 result from reallocating the 46 hours saved to new calls. These results highlight that even a conservative stopping agent, with only two decision points relatively late in the call, can yield substantial gains in sales effectiveness.

\begin{figure}[htp]
  % \vspace{-5mm}
  \centering
  \caption{Expected number of sales by our stopping agent with $T=2$ and $T=3$ decision opportunities.}
  \label{fig:expected_sales_T3}
  \includegraphics[width=\linewidth]{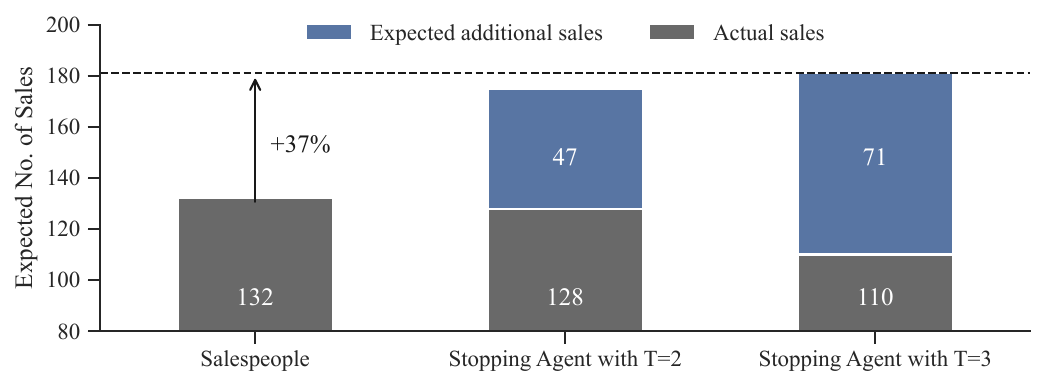}
\end{figure}

\paragraph{Extending the Decision Horizon.}
This strong performance naturally raises the question of whether quitting earlier could unlock even greater value. To explore this, we introduce a third decision point at $t=30$, retrain our stopping agent, and reevaluate its performance. As shown in Figure~\ref{fig:expected_sales_T3}, our stopping agent with $T=3$ decision opportunities at $t=30$, $t=60$, and $t=90$ achieves a 37\% gain in expected sales by reducing the time spent on calls by 69 hours (i.e., 54\%).

% More granular decision-making may be valuable when the call failure risk varies meaningfully over the course of a conversation. In our empirical setting, however, we find that quitting after $t=90$ seconds does not yield significant time savings, and that our utterance-level transcripts (i.e., not word level) are not sufficiently granular (the shortest first utterance in our test set is 60.02 seconds long).
We continue using $T = 3$ decision points at $t\in\{30,60,90\}$ for subsequent analyses, though our proposed approach can be extended to more decision opportunities.\footnote{In practice, the number of decision opportunities is limited by the granularity of the call transcription timestamps. Word-level timestamps offer the most granularity, because a quitting decision can be made after each word. However, real-time word-level transcription is more prone to errors and requires more computational resources.} We next explore heterogeneity with salesperson effectiveness and benchmark alternative stopping policies.

\paragraph{Differential Performance by Salesperson Effectiveness.}\label{sec:success_rate_heterogeneity}
We now evaluate the returns to our $T=3$ stopping agent for salespeople with different levels of effectiveness. We partition the test set into calls made by salespeople with success rates below and above the median salesperson success rate. Table~\ref{tab:expected_sales} reports the performance of salespeople and our stopping agent in each subgroup.
\begin{table}[h!]
  \vspace{1em}
  \centering
  \small
  \caption{Expected sales gain for salespeople with success rates below and above the median.}
  \label{tab:expected_sales}
  \begin{tabular*}{\linewidth}{@{\extracolsep{\fill}} lrrrr}
      \toprule  
      & \multicolumn{2}{c}{Low Success Rate Salespeople} & \multicolumn{2}{c}{High Success Rate Salespeople} \\
      \cmidrule{2-3}\cmidrule{4-5}
       & Salespeople & Stopping Agent & Salespeople & Stopping Agent\\
      \midrule
      No. of Sales (original) & 16 & 13 & 116 & 97  \\
      Total Time (hours) & 77 & 38 & 51 & 21  \\
      Additional Sales (expected) & --- & 12 & --- & 45  \\
      \midrule
      Sales Gain (\%, expected) & --- & \phantom{2s}\text{56\%}  & ---  & \phantom{2}\text{22\%} \\
      \bottomrule
  \end{tabular*}\\[0.05in]
  \vspace{-0.5em}
\end{table}

Low success rate salespeople have a success rate of 1.4\% and spend 160 seconds per call on average. High success rate salespeople have a success rate of 9.4\% and spend 224 seconds per call on average. We use these subgroup-specific averages (instead of the overall sample averages) to translate our stopping agent's time savings to the expected additional sales in each subgroup. % TODO: sensitivity analysis for each decile

Our stopping agent increases expected sales by 56\% for low-success-rate salespeople and by 22\% for high-success-rate salespeople. The gains for low-success-rate salespeople are striking because they rarely succeed and spend about 2.5 minutes per call, so any improvement in efficiency has a disproportionately large impact. These results suggest that our stopping agent not only boosts overall productivity, but may also reduce performance disparities across salespeople.
% TODO: cite?

\subsection{Evaluating Alternative Stopping Policies}\label{sec:baselines}

In this section, we compare our stopping agent against several alternative policies, ranging from simple decision rules to state-of-the-art reinforcement learning benchmarks.

\subsubsection{Simpler Decision Rules}

We first compare our stopping agent with simple policies that either ignore the conversation content or do not account for time costs.

\begin{figure*}[h!]
  \vspace{0.75em}
  \centering
  \caption{Expected number of sales by our stopping agent vs. several content and cost-agnostic baselines.}
  \label{fig:baselines_and_deadline}
  \begin{subfigure}[b]{0.33\textwidth}
    \centering
    \includegraphics[width=\linewidth]{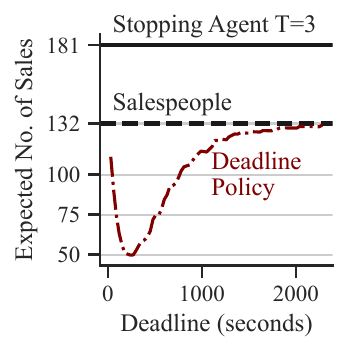}
    \caption{Expected no. of sales by our $T=3$ stopping agent and a deadline policy.}
    \label{fig:deadline}
  \end{subfigure}
  \hfill
  \begin{subfigure}[b]{0.64\textwidth}
    \centering
    \includegraphics[width=\linewidth]{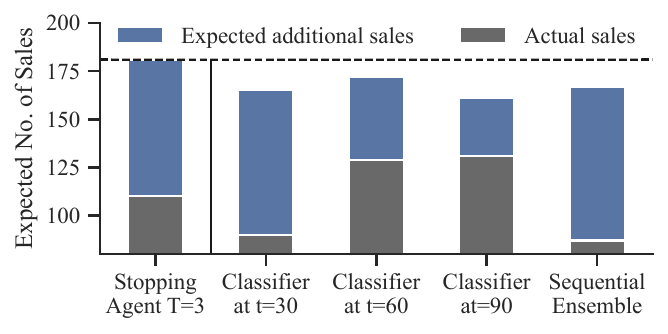}
    \caption{Expected number of sales by our $T=3$ stopping agent, classification-based policies at $t=30$, $t=60$, and $t=90$, and a sequential ensemble.}
    \label{fig:baselines}
  \end{subfigure}
  \vspace{-1.75em}
\end{figure*}

\paragraph{A deadline policy.} We begin with a basic \emph{deadline policy} that quits at a fixed time $t$ if the call has not yet succeeded by $t$. This policy does not consider call content. Figure \ref{fig:deadline} presents the expected number of sales across different deadlines, alongside the performance of both salespeople and our $T = 3$ stopping agent. Across all values of $t$, the deadline policy underperforms both salespeople and our stopping agent. This suggests that a simple rule advising salespeople to quit if a call lasts more than $t$ seconds is unlikely to improve sales effectiveness.
% This is particularly true because calls that end in a sales often take longer to close the contract.

% \vspace{-0.25em}
\paragraph{Classifier-based policies.} Next, we assess a set of classifier-based policies. Each policy consists of a GPT-4.1 language model fine-tuned to classify whether the call will eventually succeed using only the first $t=30$, $t=60$, or $t=90$ seconds of the transcript, respectively (similar to the predictor in Section \ref{sec:data}). At each $t$, the policy quits if the predicted success probability falls below a threshold selected to maximize balanced classification accuracy on the validation set. As shown in Figure~\ref{fig:baselines}, although the classifier-based policies outperform salespeople, they perform worse than our stopping agent despite being built on the same GPT-4.1 large language model.

% \vspace{-0.5em}
\paragraph{A sequential classifier ensemble.}
A possibly stronger benchmark is a sequential classifier ensemble, which combines the classifier-based policies at $t=30$, $t=60$, and $t=90$. The ensemble begins at $t=30$: if the classifier-based policy at $t = 30$ does not quit, the decision is deferred to $t=60$, and likewise to $t=90$ if needed. This gives the ensemble multiple decision opportunities like our stopping agent, but without explicitly incorporating forward-looking cumulative reward maximization. As shown in Figure~\ref{fig:baselines}, this ensemble also underperforms our stopping agent.

In summary, these results underscore the limitations of call content and time cost-agnostic stopping policies. Even when built on the same (GPT-4.1) LLM, such policies fail to match the performance of our stopping agent, which is explicitly trained to make sequential decisions using the call content that maximize the expected cumulative reward.

\subsubsection{A State-of-the-Art Reinforcement Learning Method}
We benchmark our stopping agent against a state-of-the-art deep reinforcement learning approach for optimal stopping: the optimal stopping policy gradients (OSPG) algorithm \citep{damera2023deep}.
OSPG is designed for recurrent neural network (RNN) policies, and not LLM policies.
Hence, we equip the RNN policies with linguistic knowledge by representing the transcript at $t \in \{30, 60, 90\}$ with  
3072-dimensional OpenAI \texttt{text-embedding-3-large} vectors. 
We train policies with $T=3$ decision points and report the best-performing hyperparameter configurations in Table~\ref{tab:rl_comparison}.
\begin{table}[h!]
  \vspace{0.75em}
  \centering
  \small
  \caption{Comparing our stopping agent with a state-of-the-art reinforcement learning method.}
  \label{tab:rl_comparison}
  \begin{tabular*}{\linewidth}{@{\extracolsep{\fill}} lrrrr}
      \toprule
      \multirow{2}{*}{OSPG (Reinforcement Learning), $T=3$}
      &  No. of & Total & Additional Sales  & Sales Gain \\
      &  Sales & Time (h) & (expected) &  (\%, expected) \\
      \midrule
      Salespeople & 132 & 128 & --- & --- \\
      \midrule
      Policy size = 3092 $\times$ 1, $\sim$10M parameters & 117 & 109 & 20 & \phantom{-}4\% \\
      Policy size = 3092 $\times$ 5, $\sim$50M parameters & 95 & 99 & 29 & -6\% \\
      Policy size = 3092 $\times$ 10, $\sim$100M parameters & 109 & 117 & 11 & -9\% \\
      \bottomrule
  \end{tabular*}
  \caption*{\footnotesize \textit{Note:} The policy size $H \times D$ indicates $H$ hidden units in each of $D$ hidden layers. Using the implementation by \cite{damera2023deep}, we: (i) fix $H$ to the embedding dimensionality plus 20 (i.e., $H=3072+20$) and vary $D$ to control the total number of parameters, (ii) tune the learning rate in $\{10^{-5}, 10^{-4}, 10^{-3}, 10^{-2}\}$, and (iii) train for 100 epochs using the AdamW optimizer with early stopping on the validation set.
  }
  \vspace{-1em}
\end{table}

All OSPG policies underperform our stopping agent, and the best OSPG policy achieves an expected sales gain of 4\%. In fact, the OSPG policies with $\sim$50M and $\sim$100M parameters also underperform salepeople and produce expected sales losses. We also find that many training runs fail due to the reward plateauing or collapsing to zero, consistent with the documented stability issues of deep reinforcement learning \citep{engstrom2020implementation}.

Taken together, these results highlight the limitations of both simpler heuristics that ignore the call text and time costs, and more complex reinforcement learning approaches that are not stable or scalable. They reinforce the effectiveness of imitation learning for optimizing sequential stopping decisions with high-dimensional states and externally governed state transitions. Next, we evaluate the versatility of our stopping agent by instantiating it with open-source language models.

\subsection{Parameterizing our Stopping Agent with Open-Source Language Models}
\label{sec:os_comparison}

Unlike proprietary language models, open-source language models can be hosted by firms on-premise, which allows for greater control and privacy. However, open-source language models often exhibit lower performance in terms of predictive accuracy and language understanding relative to proprietary alternatives. Given this trade-off, we next assess whether our stopping agent can still deliver efficiency gains when implemented using open-source language models.

We instantiate our stopping agent with two open-source language models: (i) Gemma 3 with 270 million parameters \citep{team2025gemma}, and (ii) Llama 3.2 with 3 billion parameters \citep{touvron2023llama}. Both models are significantly smaller than GPT-4.1 (believed to have trillions of parameters), and thus cost-effective to host on-premise. Our results are in Table~\ref{tab:os_comparison}.
\begin{table}[h!]
  \vspace{0.25em}
  \centering
  \small
  \caption{Evaluating our stopping agent with open-source language models as the policy.}
  \label{tab:os_comparison}
  \begin{tabular*}{\linewidth}{@{\extracolsep{\fill}} lrrrr}
      \toprule
      & No. of & Total & Additional Sales & Sales Gain \\
      & Sales  & Time (h) & (expected) & (\%, expected)\\
      \midrule
      Salespeople & 132 & 128 & --- & --- \\
      \midrule
      Gemma 3 (270 million parameters) & & & &\\
      \quad$T=2$ Stopping Agent & 103 & 95 & 34 & 4\% \\
      \quad$T=3$ Stopping Agent & 81 & 65 & 65 & 10\% \\
      \midrule
      Llama 3.2 (3 billion parameters) & & & &\\
      \quad$T=2$ Stopping Agent & 115 & 100 & 28 & 9\% \\
      \quad$T=3$ Stopping Agent & 86 & 63 & 65 & 16\% \\
      \bottomrule
  \end{tabular*}
  \vspace{-0.5em}
\end{table}

Even with significantly smaller open-source language models parameterizing the stopping policy, our $T=3$ stopping agent achieves expected sales gains of 10\% to 16\% over salespeople. These results show that, even under privacy and computational constraints, firms can effectively deploy our stopping agent to improve sales efficiency.

\subsection{Evaluating our Stopping Agent on an Out-of-Sample Campaign}
\label{sec:long_term_performance}

We now assess the robustness of our stopping agent to distribution shift by evaluating its performance, \textit{without retraining}, on a different outbound sales campaign conducted six months after the original campaign. In this \textit{out-of-sample} campaign, the firm targeted subscribers of a lower-cost sub-brand (analogous to the relationship between Mint Mobile and T-Mobile in the United States) with a similar offer to switch their electricity provider as in the original campaign.

The out-of-sample campaign comprises 8,334 first-contact calls made by 153 salespeople over one month. Only 31 of these salespeople (approximately 20\%)  also participated in the original campaign.
While the average time spent per call was nearly identical across campaigns (195 vs. 196 seconds), the success rate was lower in the out-of-sample campaign (4.9\%) compared to the original (5.5\%).
The difference in targeted subscribers and participating salespeople likely creates differences in the conversational content from our original campaign while maintaining the same sales objective, thereby allowing us to isolate the effect of distribution shift.

The results in Table \ref{tab:longterm_comparison} show that our stopping agent continues to improve sales efficiency in this out-of-sample campaign without retraining. Specifically, it achieves a 10\% increase in expected sales at $T=2$ and a 16\% increase at $T=3$ relative to salespeople, corresponding to 39 and 63 expected additional sales, respectively, despite the temporal gap and change in the subscriber base.

\begin{table}[h!]
  \vspace{1em}
  \centering
  \small
  \caption{Evaluating our stopping agent on an out-of-sample campaign.}
  \label{tab:longterm_comparison}
  \begin{tabular*}{\linewidth}{@{\extracolsep{\fill}} lrrrr}
      \toprule
      & No. of & Total & Additional Sales
      & Sales Gain \\
      & Sales & Time (h) & (expected) & (\%, expected)\\
      \midrule
      Salespeople & 405 & 454 & --- & --- \\
      $T=2$ Stopping Agent & 328 & 323 & 116 & 10\% \\
      $T=3$ Stopping Agent & 275 & 237 & 193 & 16\% \\
      \bottomrule
  \end{tabular*}
\end{table}

These findings suggest that the conversational patterns and decision thresholds learned by our stopping agent generalize across time and remain robust under (some) distribution shift. % Practically, this robustness reduces the need for frequent retraining.
% As a result, the stopping agent offers a scalable and field-ready solution for improving efficiency across dynamic operational contexts.

\section{Diagnosing Salespeople's Quitting Decisions}
\label{sec:diagnosing_decisions}

Having demonstrated that our stopping agent delivers substantial performance gains, we now diagnose why salespeople's quitting decisions fall short, using our stopping agent's quitting decisions as a benchmark. Since identifying salespeople's behavioral primitives directly is challenging without experimental data, we approach this question by empirically examining the behavioral patterns associated with salespeople's quitting decisions.

Specifically, in Section~\ref{sec:predicting_decisions}, we train and analyze interpretable machine learning predictors of salespeople's quitting decisions. Our analysis reveals that salespeople decide when to quit using simple decision rules that overweight a few salient phrases that often occur late in the call, whereas our stopping agent uses a more complex and dynamic set of linguistic cues. In Section~\ref{sec:opportunity_cost_shift}, we examine how salespeople's quitting behavior changes under heightened time pressure. We find evidence that salespeople do not sufficiently adjust their quitting decisions to account for call failure risk when the opportunity cost of time is higher.

Together, our results suggest that salespeople face cognitive limits when attempting to disqualify consumers early in a call, plausibly due to having to engage with the consumer, address objections, and perform other selling tasks simultaneously. This highlights the value of algorithmic support in reducing cognitive load and improving real-time disqualification decisions.

\subsection{Explaining Quitting Decisions using Interpretable Machine Learning}\label{sec:predicting_decisions}

\subsubsection{Examining quitting times}
We begin by visually examining when salespeople and our stopping agents quit calls in the test set. To do so, we use a series of Sankey diagrams that depict quitting decisions at discrete time points. To enable direct comparisons with our stopping agents, we encode the salesperson's quitting decision at $t\in\{30,60,90\}$ as a binary indicator that equals 1 if the call lasts longer than $t+\Delta$ seconds, and 0 otherwise. We set $\Delta=10$ to account for the average duration of closing salutations (though our results are robust to alternative values of $\Delta$). If a call ends between $t+\Delta$ and the next time period, we label it as ``ended'' instead of ``quit''.

\begin{figure*}[h!]
  % \vspace{1em}
  \caption{Sankey diagram of quitting decisions by salespeople and by our stopping agents.}
  \centering
  \label{fig:stopping_agent_decisions}
  \begin{subfigure}[t]{0.51\textwidth}
    \centering
    % \vspace{-53mm}
    \includegraphics[width=\linewidth]{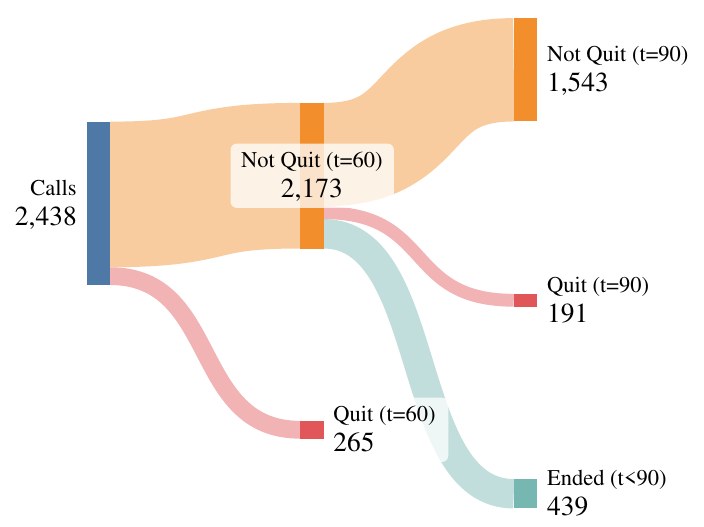}
    % Calls [2173] Not Quit (t=60)
    % Calls [265] Quit (t=60)
    % Not Quit (t=60) [839] Not Quit (t=90)
    % Not Quit (t=60) [895] Quit (t=90)
    % Not Quit (t=60) [439] Ended (t<90)
    \caption{Salespeople}
    \label{fig:sankey_salespeople}
  \end{subfigure}
  % \hfill
  \begin{subfigure}[t]{0.47\textwidth}
    \centering
    \includegraphics[width=\linewidth]{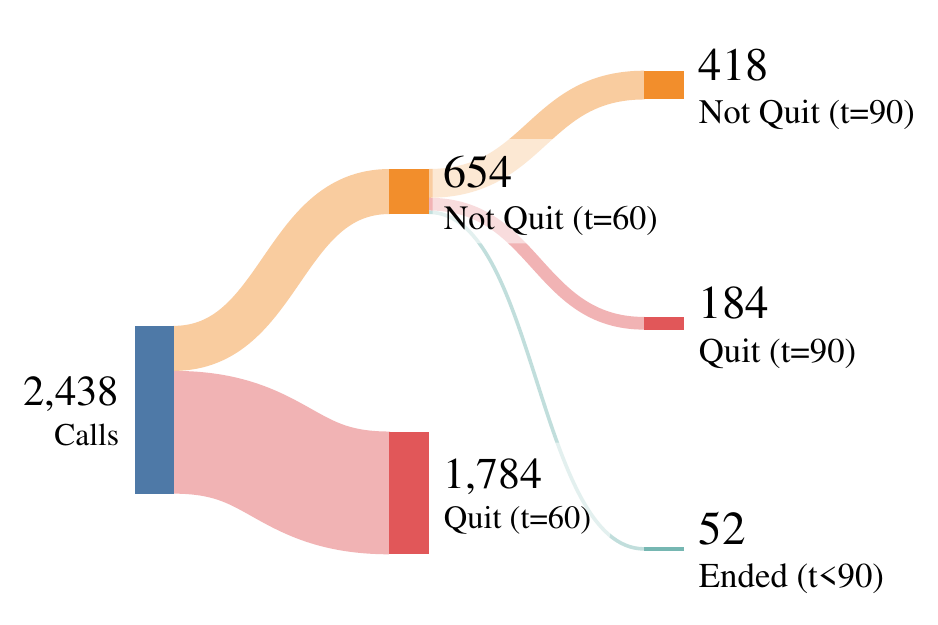}
    \caption{$T=2$ Stopping Agent}
    \label{fig:sankey_stoppingagent_T2}
  \end{subfigure}\\
  \begin{subfigure}[t]{0.47\textwidth}
    \centering
    \includegraphics[width=\linewidth]{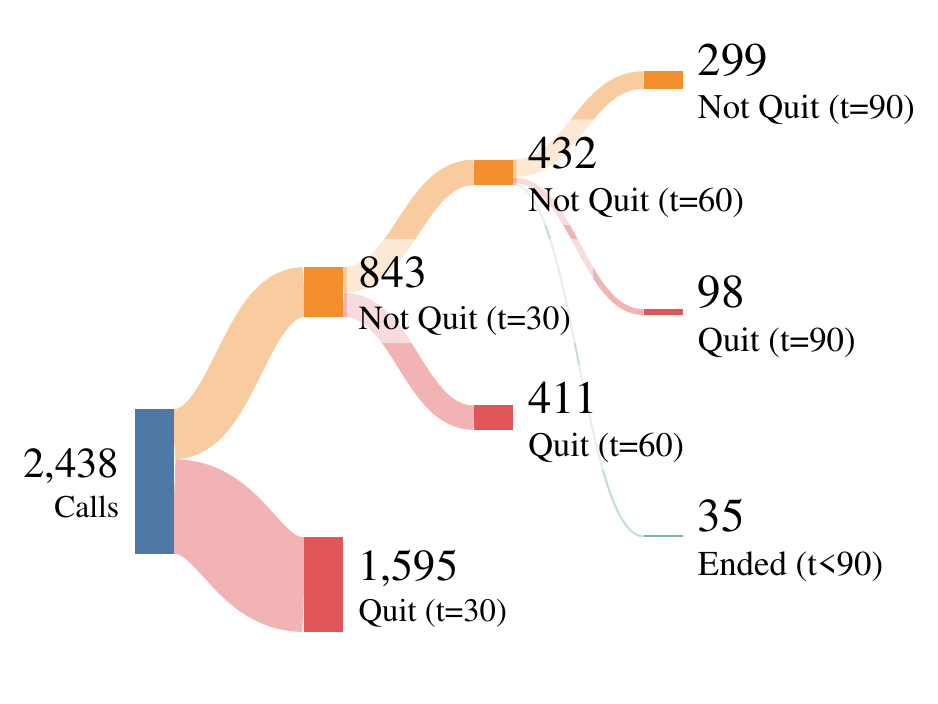}
    \caption{$T=3$ Stopping Agent}
    \label{fig:sankey_stoppingagent_T3}
  \end{subfigure}
  \caption*{\footnotesize \textit{Note:} Salespeople did not quit any calls at $t=30$, so we omit that decision opportunity from the Sankey diagram. 439 calls in the test set that were not quit by salespeople at $t=60$ (by our definition) ended before $t=90$.}
  % \vspace{-0.75em}
  % Calls [654] Not Quit (t=60)
  % Calls [1784] Quit (t=60)
  % Not Quit (t=60) [418] Not Quit (t=90)
  % Not Quit (t=60) [184] Quit (t=90)
  % Not Quit (t=60) [52] Ended (t<90)
  \label{fig:allsankeys}
  \vspace{-1.25em}
\end{figure*}

Figure~\ref{fig:allsankeys} visualizes quitting behavior using Sankey diagrams for (a) salespeople, (b) our $T=2$ stopping agent, and (c) our $T=3$ stopping agent, respectively.
Figure \ref{fig:sankey_salespeople} shows that salespeople hesitate to quit early: they quit \textit{no} calls at $t=30$, just 11\% of calls at $=60$. and just 11\% of the remaining calls $t=90$. In contrast, our $T=2$ stopping agent (Figure~\ref{fig:sankey_stoppingagent_T2}) quits 75\% of calls at $t=60$, and 31\% of the remaining calls at $t=90$. Our $T=3$ stopping agent (Figure~\ref{fig:sankey_stoppingagent_T3}) quits even more aggressively. These patterns highlight a key behavioral difference between salespeople and our stopping agents: while our stopping agents favor quitting early, salespeople hesitate and delay.

We now turn to examining \textit{why} salespeople hesitate to quit early. To answer this question, we empirically examine what linguistic signals appear to drive salespeople's quitting decisions.

\subsubsection{Examining the behavioral model underlying salespeople's quitting decisions}
To better understand the behavioral model underlying salespeople's quitting decisions, we train interpretable machine learning models to predict whether a salesperson will quit at $t=60$ using only the call text before $t=60$.
Specifically, we train random forest classifiers with varying maximum depths (to control model complexity) on the training set calls. We use the 10,000 most frequent unigrams and bigrams as input features, and following standard practice, normalize the unigram and bigram term frequencies by their inverse document frequencies (i.e., TF-IDF).
\begin{figure*}[h!]
  % \vspace{1em}
  \centering
  \caption{Predicting salespeople quitting at $t=60$ (left) and eventual call outcomes (right) given the first 60 seconds of the call transcript text using random forest models of different model complexities.}
  \label{fig:auc_quit_failure}
  \begin{subfigure}[t]{0.48\textwidth}
    \centering
    \includegraphics[width=\linewidth]{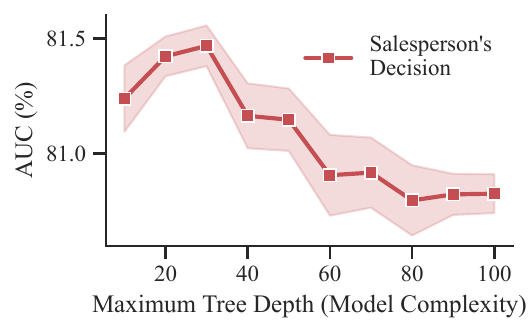}
    % \caption{}
  \end{subfigure}
  \hspace{1em}
  \begin{subfigure}[t]{0.48\textwidth}
    \centering
    % \vspace{-53mm}
    \includegraphics[width=\linewidth]{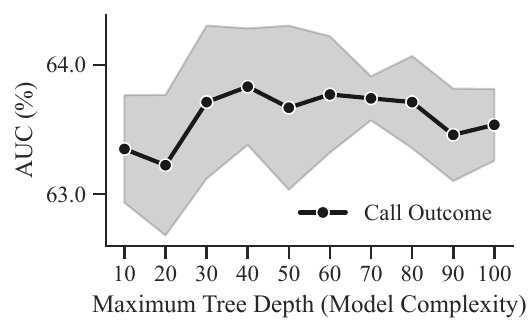}
    % \caption{}
  \end{subfigure}
  \caption*{\footnotesize \textit{Note:} Random forests with 1,000 trees are trained on unigrams and bigrams of the call transcript until $t=60$. The reported AUC is averaged over 5 runs (shaded regions depict standard deviations).}
  \vspace{-0.75em}
\end{figure*}

Figure~\ref{fig:auc_quit_failure} (left) shows the held-out AUC of predicting salespeople's quitting decisions with increasing model complexity. For comparison, Figure~\ref{fig:auc_quit_failure} (right) shows the held-out AUC of predicting eventual call outcomes given the same features and model.
These figures reveal two key insights.

First, salespeople's quitting decisions are substantially more predictable than call outcomes with random forests; the held-out AUC of predicting salespeople quitting at $t=60$ is over 17 percentage points higher. Second, the performance of predicting salespeople's quitting decisions deteriorates sharply with increasing model complexity (indicating overfitting), whereas the performance of predicting eventual call outcomes remains relatively stable with increasing model complexity.

These patterns suggest that salespeople's quitting decisions follow simple rules, which can be captured by shallow random forests but are overfit by deeper ones. In contrast, call outcomes appear to follow more complex rules that are not easily captured even by deep random forests. Since the simple rules driving salespeople's decisions may not reflect the complex rules that govern actual call outcomes, these findings suggest that salespeople are cognitively bounded. This may explain, in part, why salespeople's quitting decisions underperform relative to our stopping agent.

\subsubsection{Uncovering the linguistic cues potentially driving salespeople's quitting decisions} 
\begin{figure*}[h!]
  % \vspace{1em}
  \centering
  \caption{Most important features (by Gini importance  \citep{breiman2017classification}) in the best random forest model predicting quitting decisions at $t=60$ by (a) salespeople, and (b) our $T=2$ stopping agent.}
  \label{fig:top}
  \begin{subfigure}[t]{0.49\textwidth}
    \centering
    \includegraphics[width=\linewidth]{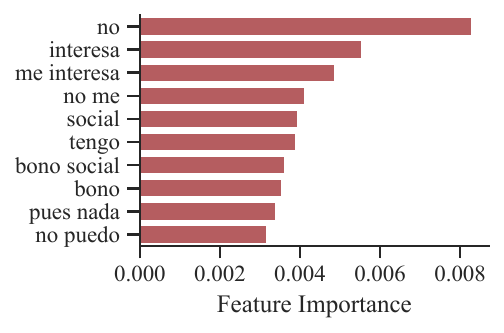}
    \caption{}
    \label{fig:top_5_quitting}
  \end{subfigure}
  \hspace{1em}
  \begin{subfigure}[t]{0.47\textwidth}
    \centering
    \includegraphics[width=\linewidth]{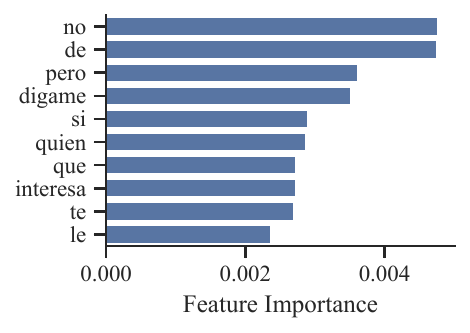}
    \caption{}
    \label{fig:top_5_quitting_stoppingagent}
  \end{subfigure}
  \vspace{-1em}
\end{figure*}

We now leverage the interpretability of random forests to uncover the linguistic cues that may drive salespeople's quitting decisions. Specifically, we select the best random forest predictor of salespeople quitting at $t=60$, and extract the most predictive features from this predictor based on their Gini importance \citep{breiman2017classification}.
Figure \ref{fig:top_5_quitting} displays the 10 most predictive features. 

Notably, the most important features predicting salespeople quitting all come from the phrase \textit{``no me interesa''} (i.e., ``I'm not interested''), suggesting that salespeople tend to quit when they hear this explicit expression of disinterest, and tend to continue the call otherwise. Further, the feature importance distribution is skewed towards the features comprising \textit{``no me interesa''}, indicating a disproportionate reliance on \textit{``no me interesa''} over other phrases.

The phrase \textrm{``I'm not interested''} holds a special place in sales: it is the canonical example of an objection in sales training materials. Its prominence likely makes it \textit{salient}, and salience is a well-documented source of bias in decision-making \citep{tversky1974judgment}. However, \textit{``no me interesa''} appears during the first 60 seconds of only 7.4\% of the test set calls that ultimately fail. If salespeople over-rely on \textit{``no me interesa''} to quit due to its salience, they may miss other less-salient indicators of failure that appear earlier in the call. Hence, waiting to hear \textit{``no me interesa''} and ignoring other indicators may explain, in part, why salespeople delay quitting.

For comparison, we show the top 10 most predictive features from the best random forest predictor of our $T=2$ stopping agent's quitting decisions at $t=60$ in Figure \ref{fig:top_5_quitting_stoppingagent}.\footnote{Note that this random forest predictor is trained to predict our $T=2$ stopping agent's quitting decisions at $t=60$ on calls in the \textit{validation set}, since our stopping agent was itself trained on calls in the training set.}. We note two observations. First, the distribution of feature importances is less skewed than in Figure \ref{fig:top_5_quitting}. Second, the most important features span several linguistic indicators of a lack of conversational progress. For example, ``no'', \textit{``pero''} (``but''), \textit{``digame''} (``tell me''), and \textit{``quien''} (``who'') indicate conversations stalled at the consumer identifying who the caller is and what they want, instead of progressing to affirmations of interest or to curiosity about the product. These phrases are more subtle indicators of disinterest than \textit{``no me interesa''}, but are prevalent early in calls that eventually failed.

Motivated by these results, we now further investigate whether salespeople systematically overweight \textit{``no me interesa''} when deciding when to quit.
We assess potential overweighting by comparing the correlation of the top 10,000 unigrams and bigrams with salespeople's quitting decisions and with the predicted call failure risk at $t=60$ (measured using GPT-4.1 as in Figure \ref{fig:risk_vs_duration}).

\begin{figure*}[!h]
  \vspace{0.5em}
  \centering
  \caption{Correlation of each of the top 10,000 unigrams and bigrams (red and blue points) with the call failure risk (predicted by fine-tuned GPT-4.1) at $t=60$ ($y$-axis) and with quitting at $t=60$ ($x$-axis) by (a) salespeople, and (b) our $T=2$ stopping agent.}
  \label{fig:ngram_correlations}
  \begin{subfigure}[t]{0.48\textwidth}
    \centering
    \includegraphics[width=\linewidth]{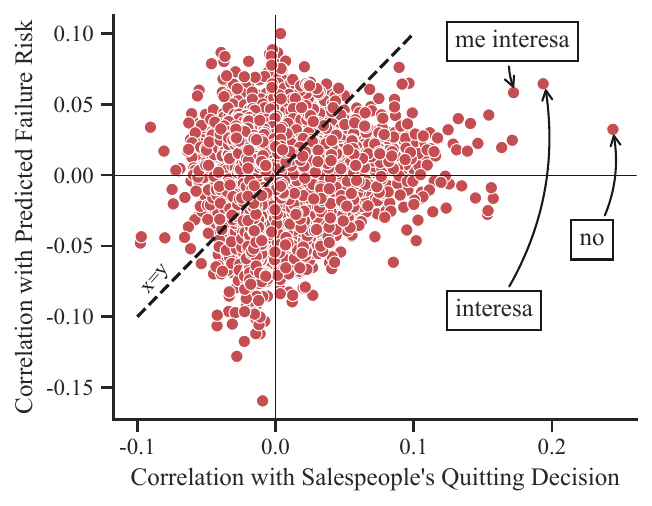}
    \caption{}
    \label{fig:ngram_correlations_salespeople}
  \end{subfigure}
  \hfill
  \begin{subfigure}[t]{0.48\textwidth}
    \centering
    \includegraphics[width=\linewidth]{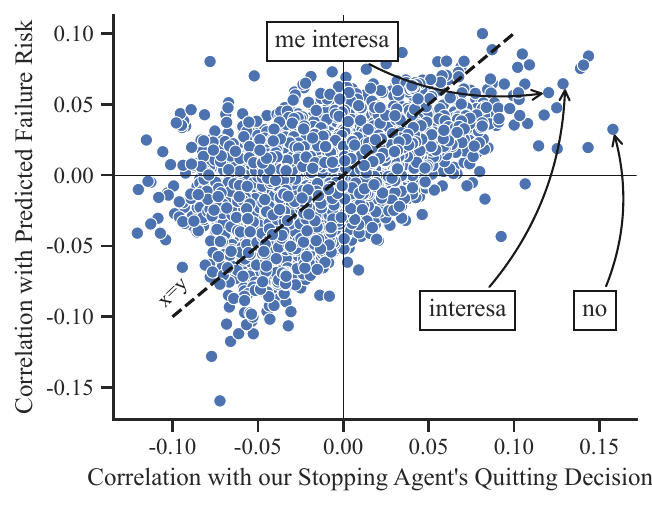}
    \caption{}
    \label{fig:ngram_correlations_stoppingagent}
  \end{subfigure}
  \vspace{-1em}
\end{figure*}

Figure~\ref{fig:ngram_correlations_salespeople} shows these correlations.
The features comprising the phrase \textit{``no me interesa''} are far more correlated with salespeople quitting ($r \in [0.17, 0.24]$) than with predicted call failure risk ($r \in [0.03, 0.06]$), and are clear outliers relative to the other unigrams and bigrams.
Hence, Figure~\ref{fig:ngram_correlations_salespeople} suggests that \textit{``no me interesa''} indeed has an outsized influence on salespeople's quitting decisions relative to other phrases and relative to its association with predicted call failure risk.

In Figure \ref{fig:ngram_correlations_stoppingagent}, we replicate Figure \ref{fig:ngram_correlations_salespeople} for our $T=2$ stopping agent. In contrast with Figure \ref{fig:ngram_correlations_salespeople}, we find that the phrases comprising \textit{```no me interesa''} do not appear to have an outsized association with quitting relative to other phrases and to their association with predicted call failure risk. Their associations are in line with the other phrases and relatively closer to the $y=x$ line than in Figure \ref{fig:ngram_correlations_salespeople}. This further suggests that our stopping agent is less biased by the salient \textit{``no me interesa''} phrase, and relies on a wider range of linguistic cues when deciding whether to quit at $t=60$.

\paragraph{Examining dynamic variation in the linguistic cues associated with quitting.} To examine how the linguistic cues associated with salespeople and our stopping agent quitting vary as the call progresses, we estimate penalized logistic regression models\footnote{We employ $L_2$ penalization to alleviate potential overfitting due to the large number of correlated independent variables in the regression (i.e., 10,000 unigrams and bigrams).} of our $T=3$ stopping agent's and the salespeople's quitting decisions on calls in the test set at each $t\in\{30, 60, 90\}$ as a function of the top 10,000 unigrams and bigrams in the call transcript text before $t$.

\begin{figure*}[!h]
    % \centering
    \small
    \vspace{0.5em}
    \caption{Top unigrams and bigrams (by logit coefficients) by decision-maker and decision opportunity.}
    \includegraphics[width=\linewidth]{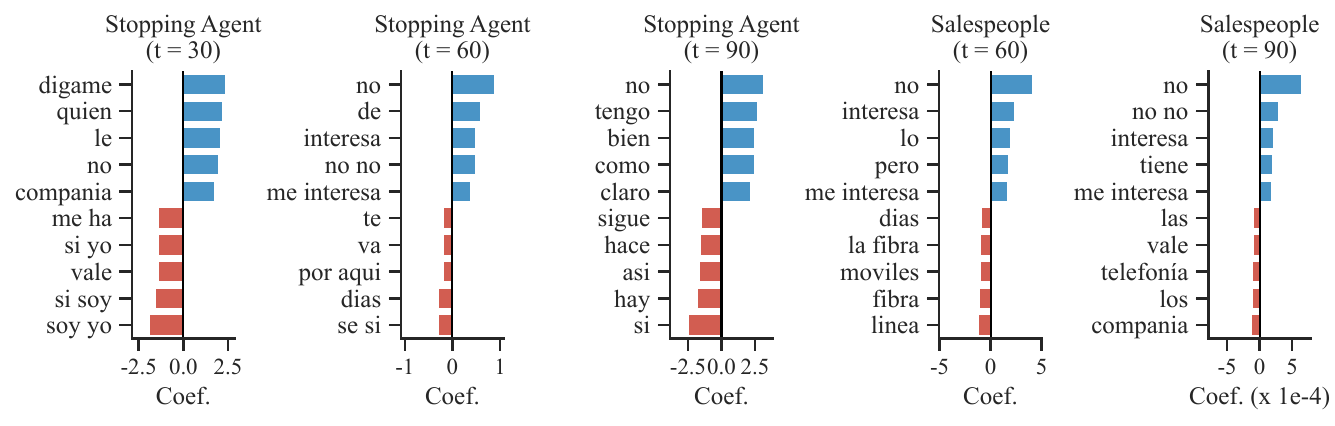}
    \caption*{\footnotesize \textit{Note:} Coefficients are from a penalized logistic regression model of the stopping agent's or salesperson's quitting decision at $t\in\{30,60,90\}$ on the top 10,000 unigrams and bigrams in the test call transcript text before $t$. The top bars indicate the 5 unigrams and bigrams with the largest positive coefficients, and the bottom bars indicate the 5 unigrams and bigrams with the largest negative coefficients. Salespeople never quit at $t=30$, so that decision opportunity is omitted. Where it appears, our partner firm's name is replaced with \textit{compania} to maintain confidentiality.}
    \label{fig:logit_words}
    \vspace{-1em}
\end{figure*}

In contrast with Figure \ref{fig:ngram_correlations}, this analysis measures \textit{conditional} correlations of each unigram and bigram with the quitting decision at $t$.
Further, it allows measuring the direction or sign of each association (unlike the feature importances in Figure \ref{fig:top}), thus complementing our previous analyses.
We report the unigrams and bigrams most positively (top) and most negatively (bottom) associated with quitting at each $t$ in Figure \ref{fig:logit_words}.

Figure \ref{fig:logit_words} reveals several insights. Consistent with Figure \ref{fig:top_5_quitting}, salespeople's quitting decisions at both $t=60$ and $t=90$ are overwhelmingly associated with the unigrams and bigrams comprising the phrase \textit{``no me interesa''} (``I'm not interested''), suggesting that salespeople continue to wait for \textit{``no me interesa''} across decision opportunities.
In contrast, the linguistic cues used by our stopping agent exhibit dynamic variation as the call progresses. Notably:

\begin{enumerate}
    \item At $t=30$, perfunctory responses such as \textit{``digame''} (``tell me'') and \textit{``quien''} (``who?'') are positively associated with our stopping agent quitting. The phrases \textit{``soy yo''} (``it's me''), \textit{``si soy''} (``yes, it's me''), and \textit{``me ha''} (``it has''), however, signal affirmation and are negatively associated with our stopping agent quitting. These patterns suggest that, at $t=30$, our stopping agent quits based on whether the salesperson is speaking to the right person.

    \item At $t=60$, the phrases positively associated with our stopping agent quitting shift to explicit indicators of disinterest, such as \textit{``no''} and \textit{``interesa''} (``interested''), while the negative associations now come from phrases that typically precede volunteering or asking for more information, such as \textit{``se si''} (``if'') and \textit{``por aqui''} (``here''). Thus, between $t=30$ and $t=60$, our stopping agent moves from quitting based on identifying \textit{who} the salesperson is talking to, to assessing \textit{what} the consumer's degree of interest or disinterest is.

    \item At $t=90$, only 397 calls remain for our stopping agent to quit: the others are either quit by our stopping agent or by a salesperson before $t=90$. The phrases positively associated with our stopping agent quitting at $t=90$ include refusal (\textit{``no''}), and constraint justifications such as \textit{``tengo''} (``I have''), whereas the phrases most negatively associated with our stopping agent quitting all indicate conversational progress. Hence, at $t=90$ the stopping agent quits based on the \textit{why} the prospect may not want the salesperson's offering. 
\end{enumerate}

Overall, these results suggest that salespeople not only rely on a limited set of linguistic cues to decide whether and when to quit, but also do not update the cues they rely on as the conversation progresses. This is consistent with them being cognitively bounded. In contrast, our stopping agent uses a wide variety of linguistic cues, and exhibits dynamic variation in the linguistic cues it uses, to decide whether and when to quit.

\subsection{Assessing Quitting Behavior Under Different Opportunity Costs of Time}\label{sec:opportunity_cost_shift}

We previously used interpretable machine learning to diagnose deficiencies in salespeople's quitting decisions. Previous research shows that decision-making deficiencies are more pronounced during periods of high fatigue or pressure (e.g., \cite{danziger2011extraneous} find that judges are less likely to grant parole near their meal break). Motivated by this research, we now analyze whether and how salespeople adjust their quitting behavior under higher time pressure or opportunity costs of time.

Intuitively, when time becomes more costly, we would expect salespeople to reallocate effort toward higher-value opportunities. In particular, they should be more likely to shorten calls with a high risk of failure, and preserve time for those with a low risk of failure. Such behavior would indicate risk-sensitive time management. However, if salespeople do not shorten high-risk calls more than low-risk ones when time is scarce, it suggests that salespeople mispredict whether a call will fail or are insensitive to call failure risk.

\paragraph{Operationalizing variation in opportunity costs of time.}
To operationalize variation in opportunity costs of time, we use whether a call was made near the end of the salesperson's shift as a cost-shifter. Calls made near the end of the shift may have higher opportunity costs due to salespeople eager to finish on time, avoid starting long conversations, or meet their daily quotas.

\begin{figure*}[h!]
  \vspace{1em}
  \centering
  \caption{Whether a call was made near the end of the salesperson's shift as a time cost-shifter.}
  \label{fig:costshifter}
  \begin{subfigure}[b]{0.47\textwidth}
    \centering
    \includegraphics[width=\linewidth]{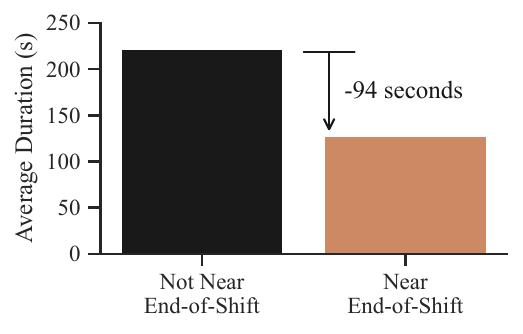}
    \caption{Avg. duration of ``near end-of-shift'' calls vs. others.}\label{fig:avg_duration_by_proximity}
  \end{subfigure}
  \quad
  \begin{subfigure}[b]{0.47\textwidth}
    \centering
    \includegraphics[width=\linewidth]{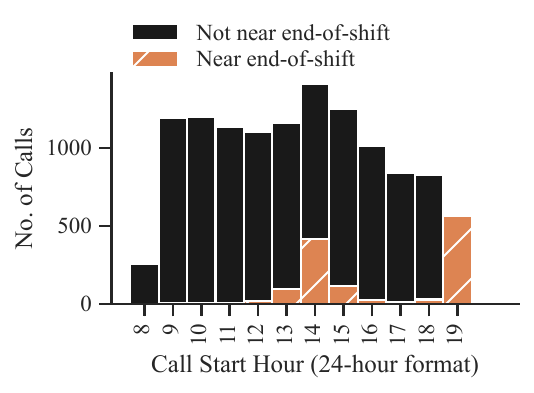}
    \caption{Start times of ``near end-of-shift`` calls vs. others.}\label{fig:call_start_hour}
  \end{subfigure}
  \caption*{\footnotesize \textit{Note:} The left plot shows the average duration of the ``near end-of-shift'' (i.e., in the first time-till-end-of-shift decile) calls and the other calls. The right plot shows the start time (hour) of the ``near end-of-shift'' calls and other calls. }
  \vspace{-2em}
\end{figure*}

We compute the time between the start of each call and the end of the salesperson's shift, and bin test set calls into deciles based on their time-till-end-of-shift.  We denote calls in the first decile as ``near end-of-shift''.
Figure~\ref{fig:avg_duration_by_proximity} shows that ``near end-of-shift'' calls are 94 seconds shorter on average than earlier calls ($p < 0.001$), suggesting that salespeople indeed face higher opportunity costs of time during these calls.
Figure~\ref{fig:call_start_hour} shows the calls' start times and confirms that ``near end-of-shift calls'' are made in the last hour of salespeople's shifts, which end at 3 p.m. and 8 p.m..

\begin{figure*}[h!]
  \vspace{1em}
  \centering
  \caption{How salespeople allocate time across calls in the test set with varying predicted failure risks, comparing calls with a low (not near end of shift) vs. a high (near end of shift) opportunity cost of time.}
  \label{fig:costshifterresponse}
  \begin{subfigure}[b]{0.44\textwidth}
    \centering
    \includegraphics[width=\linewidth]{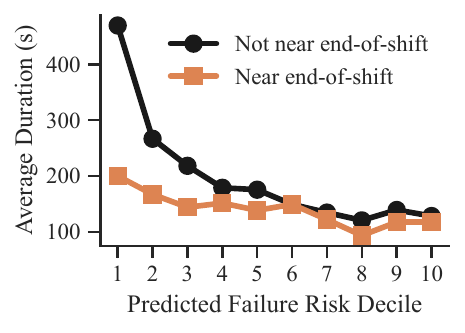}
    \caption{Call duration by predicted failure risk}
    \label{fig:duration_by_risk}
  \end{subfigure}
  \hspace{1em}
  \begin{subfigure}[b]{0.44\textwidth}
    \centering
    \includegraphics[width=\linewidth]{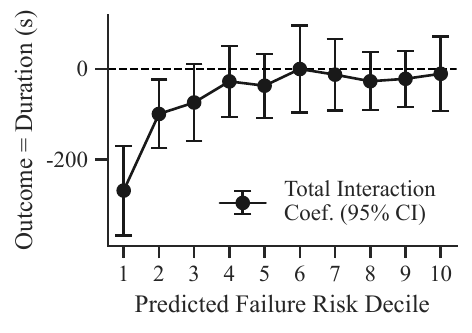}
    \caption{Total interaction effect coefficients}
    \label{fig:duration_difference_by_risk}
  \end{subfigure}
    \caption*{\footnotesize \textit{Note:} The predicted failure risk of a call is the probability of the call failing to end in a sale, binned into deciles, predicted by a fine-tuned GPT-4.1 model given the first 60 seconds of the call transcript. The right plot shows the total interaction effect coefficients and 95\% confidence intervals from a regression of the call duration on the predicted failure risk decile interacted with a dummy for near end-of-shift calls (i.e., first time-till-end-of-shift decile).}
  \vspace{-1em}
\end{figure*}

\paragraph{Do salespeople allocate call time in a risk-sensitive manner?}
We now examine whether salespeople respond to higher opportunity costs of time by reallocating time more effectively across calls with different predicted failure risks (measured using GPT-4.1 as in Figure \ref{fig:risk_vs_duration}). Figure~\ref{fig:duration_by_risk} plots the average call duration by predicted failure risk decile (for test set calls), separately for near end-of-shift calls and for calls made earlier in the shift. If salespeople allocate time in a risk-sensitive manner, we would expect them to (1) spend lesser time on high-risk calls, and (2) differentially reduce the time spent on high-risk calls when the opportunity cost of time is higher.

Figure~\ref{fig:duration_by_risk} shows that for calls made earlier in the shift, salespeople do spend less time on higher risk calls (i.e., the curve slopes downwards). However, near the end of their shift, salespeople shorten calls \textit{throughout} the predicted failure risk distribution. In fact, they reduce time the most on calls in the lowest-risk deciles (1 and 2); calls that, ex-ante, were more likely to succeed.

To formally test this pattern, we estimate the following regression:
\begin{align}
  \textrm{Duration}_i  = \alpha_0 + \gamma_0 s_i + \sum_{j=2}^{10} \alpha_j p_{ij} + \sum_{j=2}^{10} \beta_j p_{ij}s_i + 
   \epsilon_i, \label{eq:costshifterresponse}
\end{align}
where $p_{ij}$ is an indicator that call $i$ falls in predicted failure risk decile $j$, and $s_i$ is an indicator for the call being made near the end of the salesperson's shift. The coefficients $\beta_j$ capture how the effect of being near the end of a shift varies across risk deciles.
%where $p_{ij}=\mathbb{I}[\textrm{Predicted failure risk decile of call }i=j]$ and $s_i =\mathbb{I}[\textrm{Call } i \textrm{ made near end-of-shift}]$.

Figure~\ref{fig:duration_difference_by_risk} shows the estimated total interaction coefficients (i.e., $\gamma_0$ for decile 1 and $\gamma_0 + \beta_j$ for deciles 2 to 10) and their 95\% confidence intervals. 
We find statistically significant time reductions in low-failure-risk deciles (1 and 2) ($p<0.001$), but no systematic shortening of calls in the other deciles. This pattern suggests that salespeople are not responding to higher time costs by reducing the time spent on high-risk calls, but are instead shortening calls across the board, or even disproportionately shortening calls that were more promising ex-ante.

These results, taken together with our earlier findings, point to a broader failure to integrate risk information into disqualification decisions, even when opportunity costs are elevated. This behavior may stem from salespeople's limited ability to accurately assess call failure risk.\footnote{In Appendix~A, we supplement this analysis with a contraction test \citep{kleinberg2018human}, which examines whether salespeople misrank calls by their failure risk under the assumption that selling is purely informative. The contraction test compares salespeople with an imperfect algorithm that predicts call outcomes at $t=60$, and reveals that 80\% of the salespeople in our dataset misrank calls by their failure risk.}

\subsection{Summary and Discussion}
Our findings suggest that salespeople rely heavily on simple functions of a few salient cues (most notably, the phrase ``no me interesa'') that often appear late in the call, do not adjust the cues they rely on as the call progresses, and do not adjust their quitting behavior in response to elevated opportunity costs of time. When faced with higher opportunity costs of time near the end of their shift, salespeople shorten calls indiscriminately, including those with strong sales potential.

Together, these patterns suggest that suboptimal quitting arises due to cognitive bounds: a limited ability to integrate failure risk, time costs, and conversational context into real-time decisions. This reinforces the need for decision-support tools like our stopping agents that systematically incorporate these factors to guide real-time quitting decisions.

\section{Conclusion}\label{sec:closing}

% summary of what we have done
In high-volume outbound sales settings, leads are ample yet most calls fail to end in a sale and consume substantial time and resources. This inefficiency poses a core managerial challenge: how to identify, in real time, which conversations are worth continuing and which should be abandoned to pursue other leads. We address this problem by introducing stopping agents: generative language agents trained via imitation learning to make sequential quit or wait decisions given the evolving transcript of each call to maximize the expected cumulative reward.

% results recap
Our stopping agent delivers substantial gains. When applied to real-world sales conversations, our stopping agent reduces the time spent on failed calls by 54\% while preserving nearly all sales. Reallocating the time saved to new calls increases expected sales by up to 37\%. Applying our stopping agent to an out-of-sample campaign that was not used for training also produces a substantial expected sales gain of 16\%, suggesting that the performance of our stopping agent is durable and robust to delayed retraining.

Notably, our stopping agent delivers these gains using only the transcript text (without visuals or voice), and can be trained and deployed cost-effectively using either commercial APIs or open-source models. As such, our stopping agent offers a practical decision support tool that integrates with existing real-time call transcription workflows. We release open-source code that allows companies to implement our stopping agent on their own data.

Beyond performance improvements, our analysis offers insight into the behavioral foundations of salesperson inefficiency. We show that salespeople seem to rely on salient cues (e.g., ``\textit{no me interesa}'') that often appear late in the call, in contrast with our stopping agent that relies on subtle linguistic cues that appear earlier in the call, and that exhibits dynamic variation in the cues it relies on. These patterns suggest that cognitive bounds are a key constraint in dynamic qualification decisions, and that algorithmic decision support may help reduce the impact of these bounds.

\paragraph{Future work.} This work opens several promising directions for future research. In this paper, we focus on short-term sales.\footnote{Discussions with our partner firm indicate that short-term sales was indeed the goal of the campaigns we study.} Theoretically, one could easily adapt the reward structure to long-run objectives such as customer value or brand outcomes, which may encourage agents to persist longer in conversations towards achieving these objectives. 

Another extension involves expanding the action space of our stopping agent. Given the versatility of the underlying language models, our approach could be extended to support a broader set of conversational decisions beyond \textsf{quit} and \textsf{wait}, for example, recommending scripted responses (à la Google's smart-reply \citep{kannan2016smart}), asking clarifying questions, or suggesting pauses. Such multi-action policies would allow the agent to not only determine whether to quit, but also how to steer the conversation more effectively. However, this extension would require additional data from interactions between our stopping agent and salespeople (and the consumer's responses), a more complex estimation procedure, and an off-policy evaluation protocol.
%One is to combine upstream lead-scoring models with downstream stopping agents to optimize the full engagement pipeline.

Our work focuses on the textual information of the call to guide the stopping agent's decisions. Hence, our approach relies on high-quality transcripts and may be sensitive to inaccuracies in automatic speech recognition, transcription, and diarization. Foundation models for voice and speech are rapidly evolving (e.g., \citep{baevski2020wav2vec}). Future research could leverage these models to directly incorporate voice and non-verbal signals (e.g., tone, pitch, and prosody) to enhance our stopping agents. 
Finally, understanding how salespeople respond to AI-generated recommendations remains an open managerial and behavioral question that could be addressed via field experimentation \citep{kawaguchi2021will,dietvorst2018overcoming}.

\paragraph{Limitations.}
While our findings demonstrate the potential of stopping agents to improve salesforce productivity, several limitations warrant consideration. Our empirical setting involves one firm, one product category, and one language. While the stopping agent generalizes well within this context, its applicability to other industries, particularly those involving persuasive, consultative, or relationship-based selling, remains to be tested. Research on benchmarking and improving the persuasive ability of LLMs is nascent \citep{jin2024persuading,pauli-etal-2025-measuring}, and how to generate persuasive language to optimize a long-term objective remains an open problem.

Our stopping agent is designed to save time while minimizing lost sales when salespeople spend too long on calls. Our empirical evidence from sales calls in the field demonstrates that spending too long on calls that eventually fail is indeed prevalent. However, our stopping agent is not designed to advise salespeople to persist in conversations that they decide to quit. Addressing this limitation is possible using offline-to-online reinforcement learning, where offline imitation learning is followed by online policy improvement through direct interactions with salespeople \citep{yue2024ollie} in a field experiment. We leave this endeavour to future research.

Our stopping agent also inherits the limitations of the \textit{reward hypothesis} underpinning all of reinforcement learning \citep{sutton1998reinforcement}, that ``all of
what we mean by goals and purposes can be well thought
of as maximization of the expected value of the cumulative
sum of a received scalar signal (reward)''. Recent research has infused reinforcement learning algorithms with economic structure such as downward-sloping demand curves \citep{misra2019dynamic} and intertemporal budget constraints \citep{ko2024target}. We demonstrate substantial gains without assuming such economic structure, and delegate exploration along this dimension to future work.

\paragraph{Summary.}
In sum, we show that the decision of whether and when to quit a sales conversation is not merely an intuitive judgment. Rather, it is a high-stakes optimization problem that can be solved effectively with AI. By surfacing systematic human errors and offering a scalable remedy, stopping agents transform conversational data from passive record to actionable input, paving the way for more intelligent, real-time decision support.

More broadly, our findings contribute to ongoing conversations in marketing and AI research. They illustrate how high-dimensional, unstructured data, such as live conversation transcripts, can be converted into actionable decision support through imitation learning. They also underscore the growing feasibility of embedding AI agents into frontline operational contexts, not as replacements for human workers, but as silent companions that improve outcomes in real time. Our approach offers a model for building behaviorally-informed AI systems that enhance decision quality while revealing the cognitive limits of human judgment.

\begin{small}
\singlespacing
\noindent\textrm{\textbf{Declarations.} \textit{Funding and Competing Interests:} All authors certify that they have no affiliations with or involvement in any organization or entity with any financial interest or non-financial interest in the subject matter or materials discussed in this manuscript. The authors acknowledge financial support from OpenAI.
}
\end{small}

\singlespacing

\bibliographystyle{emaad}
\bibliography{scibib}
%\printbibliography
% \end{small}

\clearpage

\doublespacing 

\renewcommand{\thesection}{A}

\section*{Appendix A: A Contraction Test of Salespeople's Quitting Decisions}
\label{app:contraction}

\cite{kleinberg2018human} introduce \emph{contraction} as a nonparametric empirical test to detect potential misranking by human decision-makers in settings where counterfactual outcomes are observed for some but not all decisions. For example, whether a defendent commits a crime before trial is observable if the judge releases them, but not if they are detained.
%TODO: Explore this route

The key insight behind the test is that even when we cannot observe the outcomes of some decisions, we can assess humans' ranking quality by comparing their decisions to those that would have been made by a predictive algorithm. The contraction test simulates replacing human decisions with algorithmic ones in order of the algorithm's predicted risk or value—\emph{contracting} the set of positive human decisions. For example, when testing judges for misranking, \cite{kleinberg2018human} simulate the release of defendants in order of their algorithm-predicted risk of reoffending.

This test relies on the assumption that the decision does not itself affect the counterfactual outcome, only who is selected into the decision. For example, in \cite{kleinberg2018human}, a judge releasing or detaining a defendant does not change the defendant's underlying propensity to commit a crime; the decision to release only \emph{reveals} but does not affect the crime risk. Under this assumption, one can compare outcomes across different decision-makers with varying thresholds (e.g., stricter versus more lenient judges) to assess whether the ranking of released individuals implied by judges decisions is consistent with outcome risk.

We apply the contraction test to quitting decisions at time $t$, analogous to a judge's decision to jail a defendant. We assume that the salesperson's decision to not quit the call at $t$ has no causal effect on the likelihood of a sale, it simply allows the outcome to reveal itself. This assumption allows us to treat variation in salespeople's behavior (e.g., some salespeople persist more than others) as a source of quasi-random variation in treatment assignment, and thus to compare outcomes across salespeople with different levels of persistence. Applying the contraction test in this setting allows us to assess whether salespeople are effectively identifying and prioritizing high-value leads, or whether an algorithm trained to predict conversion would produce better-ranked decisions.

\begin{figure*}[h!]
  \vspace{1em}
  \centering
  \caption{Quitting rate of each salesperson at $t=30$, $t=60$, and $t=90$ seconds.}
  \label{fig:contraction_quitting_rate}
    \centering
    \includegraphics[width=0.45\linewidth]{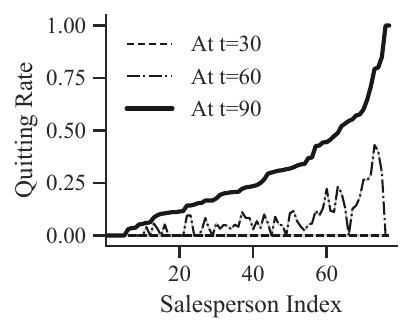}
  \end{figure*}

 We operationalize quitting at $t$ as a binary indicator of whether the call ends before $t+\Delta$ seconds, where $\Delta=10$ reflects the average duration of closing salutations. The contraction test requires variation in quitting behavior across individuals. Figure~\ref{fig:contraction_quitting_rate} shows each salesperson's quitting rate at $t=30$, $t=60$, and $t=90$ seconds. Because quitting rates exhibit limited variation at $t=30$ and $t=60$, we focus on $t=90$, where there is sufficient heterogeneity to apply the test.

\begin{figure*}[h!]
  \vspace{1em}
  \centering
  \caption{Testing for salespeople misranking calls via contraction \citep{kleinberg2018human}}
  \label{fig:contraction}
    \centering
    \includegraphics[width=0.5\linewidth]{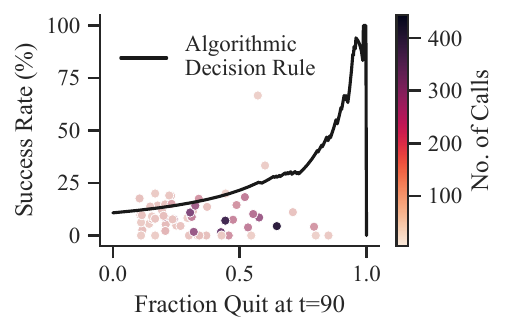}
  \caption*{\footnotesize \textit{Note:} The algorithmic decision rule curve is constructed by simulating quitting calls in the test set made by salespeople with quitting rates in the first quintile (analogous to lenient judges in \citep{kleinberg2018human}) after $t=90$ seconds, in decreasing order of calls' failure probability predicted by a fine-tuned GPT-4.1 model given the call transcript at $t=90$. Each point is a non-lenient salesperson with a quitting rate in quintiles 2 to 5.}
  % \vspace{-5mm}
\end{figure*}

To construct the algorithmic decision rule, we simulate quitting calls from the test set made by the most lenient salespeople, those in the first quintile of the quitting rates distribution at $t=90$, in decreasing order of predicted failure risk. Predictions are generated by a fine-tuned GPT-4.1 model given the call transcript up to $t=90$.
Figure \ref{fig:contraction} plots the resulting algorithmic decision rule curve. At a quitting rate of 0\%, the curve reproduces the observed success rate of lenient salespeople (10.8\%). At 100\%, the success rate falls to 0\%. Between these extremes, the success rate of the algorithmic decision rule increases as predictably-risky calls are removed, though not monotonically since the algorithm is imperfect and may misrank calls.

The contraction test then compares this curve to human decision-making. For each non-lenient salesperson (quintiles 2-5), we compute their observed success rate $y_\textrm{human}$ for calls that were not quit at $t=90$ and compare it to the algorithm's counterfactual success rate $y_{\textrm{algorithm}}(q)$ at the same quitting rate $q$. If $y_{\textrm{human}} < y_{\textrm{algorithm}}(q)$, it implies that the salesperson could have achieved higher success rate by adopting the algorithm's ranking. In other words, the salesperson is misranking calls by their failure risk.

For this comparison to be valid, we must assume that all salespeople draw from the same distribution of calls. In our setting, this assumption is plausible: prospect lists are randomly assigned to salespeople, and we find no evidence of systematic differences in the distribution of predicted success probabilities of calls across salespeople.

Figure~\ref{fig:contraction} overlays the observed performance of non-lenient salespeople (in the test set) as points on the algorithmic decision rule curve. We find that 48 of 60 non-lenient salespeople (80\%) lie below the algorithmic decision rule curve (i.e., with $y_\textrm{human} < y_\textrm{algorithm}(q)$), indicating that their ranking of calls is outperformed by the algorithm's in terms of the success rate. This supports our earlier finding that salespeople mispredict call outcomes and highlights the value of decision support tools that incorporate predictive signals more systematically.

In sum, the contraction test provides independent validation that many salespeople misrank calls by their failure risk, even when given more time to gather information. This evidence strengthens our main claim: suboptimal quitting behavior reflects bounded rationality in risk assessment—reinforcing the need for decision support tools like stopping agents.

\end{document}